%% file: main.tex
\documentclass{article}

\usepackage[final]{neurips_2025}

\usepackage[utf8]{inputenc} 
\usepackage[T1]{fontenc}    
\usepackage{hyperref}       
\usepackage{url}            
\usepackage{booktabs}       
\usepackage{amsfonts}       
\usepackage{nicefrac}       
\usepackage{microtype}      
\usepackage{xcolor}         
\usepackage{tabularx} 
\usepackage{graphicx}
\usepackage{algpseudocode}
\usepackage{wrapfig}
\usepackage{tcolorbox}
\usepackage{algorithmicx}

\algtext*{EndWhile}
\algtext*{EndFor}
\algtext*{EndIf}
\input{preamble}

\title{Images are Worth Variable Length of Representations}
\begin{document}

\author{%
Lingjun Mao$^{1}$ \hspace{0.5cm}
Rodolfo Corona$^{2}$ \hspace{0.5cm}
Xin Liang$^{2}$ \hspace{0.5cm}
Wenhao Yan$^{3}$ \hspace{0.5cm}
Zineng Tang$^{2\dagger}$ 
\\[0.3em]
$^1$University of California, San Diego\hspace{0.3cm}
$^2$University of California, Berkeley \\
$^3$University of Washington 
\\[0.3em]
\texttt{lingjun@ucsd.edu, \{rcorona, terran, xinl\}@berkeley.edu} \\
\texttt{wenhao77@uw.edu}
\\[0.3em]
$^\dagger$Corresponding author
}

\maketitle

\input{sec/0_abstract}

\input{sec/1_intro}
\input{sec/3_method}

\input{sec/4_exp}

\input{sec/2_related_works}
\input{sec/5_conclusion}
{
    \small
    \bibliographystyle{plain}
    \bibliography{main}
}

\newpage
\appendix
\input{Appendix/Implementation}
\input{Appendix/VLM}
\input{Appendix/Linear_Prob}

\input{Appendix/Quantize}

\end{document}

%% file: preamble.tex
\usepackage{amsmath,amssymb}
\usepackage{multirow}
\usepackage{subcaption}


%% file: sec/0_abstract.tex
\vspace{-15pt}
\begin{abstract}
Most existing vision encoders map images into a fixed-length sequence of tokens, overlooking the fact that different images contain varying amounts of information. For example, a visually complex image (e.g., a cluttered room) inherently carries more information and thus deserves more tokens than a simple image (e.g., a blank wall). To address this inefficiency, we propose DOVE, a dynamic vision encoder that produces a variable number of visual tokens (i.e., continuous representation vectors) to reconstruct each image. Our results show that DOVE significantly reduces the average number of tokens while maintaining high reconstruction quality. In several linear probing and downstream multimodal tasks, it outperforms existing autoencoder-based tokenization methods when using far fewer tokens, capturing more expressive semantic features compared to fixed-length encoding. We further extend DOVE with query-conditioned tokenization. By guiding the model to focus on query-relevant regions, it achieves more efficient and targeted semantic extraction. Our code and checkpoints are available at \url{https://dove-encoder.github.io/dove-encoder}.

\end{abstract}

\vspace{-10pt}
\begin{figure}[!h]
    \centering
    \includegraphics[width=0.85\linewidth]{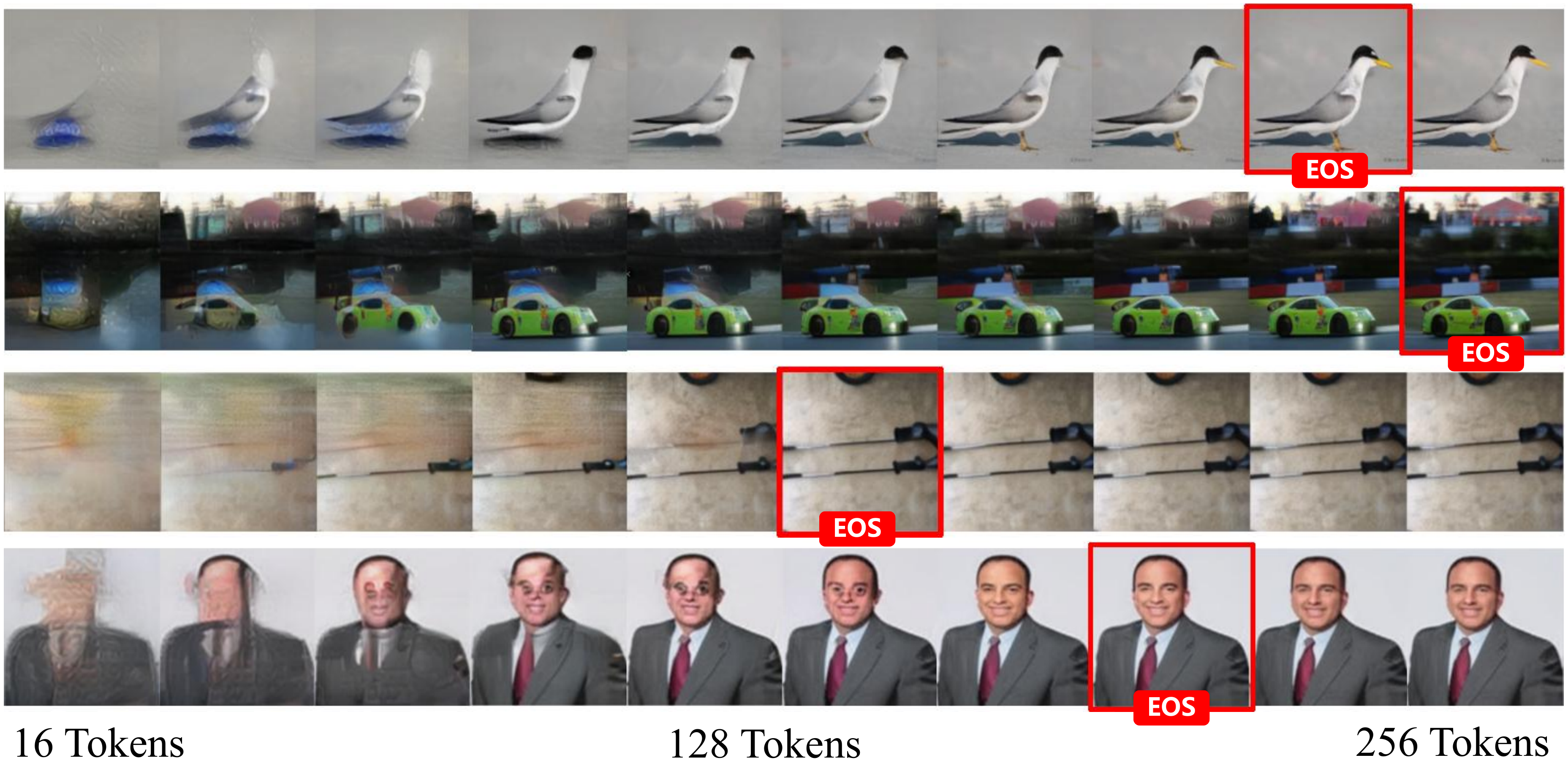}
    \caption{\textbf{Dynamic Visual Representations.} As the number of tokens used by DOVE increases, the reconstructed images shows finer and high frequency details.}
    \label{fig:DOVE_Results}
    \vspace{-10pt}
\end{figure}

%% file: sec/1_intro.tex
\section{Introduction}
Image representation learning~\cite{xia2014supervised} is a fundamental component of computer vision; it plays a pivotal role in various visual tasks, including image classification~\cite{lu2007survey, chen2021review}, object detection~\cite{zou2023object, zhao2019object}, and semantic segmentation~\cite{guo2018review, hao2020brief}. Vision representation models are also widely used in multi-modal learning, where they serve as powerful vision encoders within vision-language models (VLMs), converting image information into discrete token sequences. Existing image representation learning methods generally fall into two categories: semantic feature learning (e.g., CLIP~\cite{radford2021learning}, DINO~\cite{caron2021emerging}) and autoencoder-based image tokenization (e.g., VQGAN~\cite{esser2021taming}, VAE~\cite{kingma2013auto}). All of which aim to generate fixed length sequences. However, studies have shown that vision tokens suffer from information redundancy~\cite{chen2024efficient}. We conjecture that different images have different complexity such that they can be represented with different lengths of tokens for reconstruction.

To this end, we propose DOVE (\underline{D}ynamic \underline{O}utput \underline{V}ision \underline{E}ncoder), a visual tokenizer that adaptively generates variable-length sequences of continuous visual tokens for image reconstruction. Our method extends the standard visual autoencoder framework by incorporating a transformer-based dynamic token generator (Figure~\ref{fig:DOVE_Framework}), which is capable of generating an end-of-sequence (EOS) token at any position to terminate the output sequence. We jointly optimize image reconstruction quality and EOS token prediction based on an MSE threshold, and truncate token sequences at the predicted EOS. Our method effectively shortens the token sequence length while maintaining high reconstruction quality (Figure~\ref{fig:DOVE_Results}). As token sequences progress, their reconstructions show more high-frequency details and additions of objects, and then saturate at (EOS) token. 

By learning dynamic token lengths, we find that the tokenizer learns richer semantics and observe the emergence of zero-shot semantic segmentation by PCA on the hidden features.
We perform extensive experiments on reconstruction, classification, and question answering by replacing vision backbones in vision language models. Our approach consistently and significantly outperforms other autoencoder-based tokenization methods while enjoying improved efficiency from dynamic length. 

Considering that human vision is an active and task-driven process, and that humans tend to focus on task-relevant regions while ignoring irrelevant ones when answering questions~\cite{bajcsy2018revisiting,land1999roles,deangelus2009top}, we additionally introduce a query-conditioned variant of DOVE. This model is able to read the user's query and reconstruct the input by focusing on semantically relevant regions, thereby further reducing the length of the generated token sequence. In practice, given a text query and a corresponding salient image region during training, we feed the text query to the token generator and apply higher weights to the reconstruction loss specifically corresponding to the salient region. We find that this approach further improves token efficiency, semantics, and vision language model performance. 

We summarize our contributions as follows:
\begin{itemize}
\vspace{-5pt}
\item We propose DOVE, a visual tokenizer that dynamically generates tokens based on image complexity. Unlike previous visual tokenization, our model supports arbitrary control over the token sequence length in a single parallel forward.
\item We propose a variant of DOVE that grounds token generation on a text query and its corresponding salient visual regions. This query-conditioned model achieves a higher token compression rate (averaging 68\%) and demonstrates stronger semantic representation.
\item We observe a phenomenon of emergent semantics by probing the latent representation. Compared to other autoencoder-based tokenization methods with fixed-length token representations, our model achieves significantly better performance on classification, vision-language QA, and shows emerging semantic segmentation properties.
\end{itemize}
\vspace{-10pt}

%% file: sec/3_method.tex
\section{Dynamic Vision Tokenizer}


We introduce DOVE, a dynamic vision encoder that adaptively generates a variable number of continuous visual tokens to reconstruct each image.
\vspace{-5pt}

\subsection{Model Architecture}

\begin{figure}[!h]
    \centering
    \includegraphics[width=0.95\linewidth]{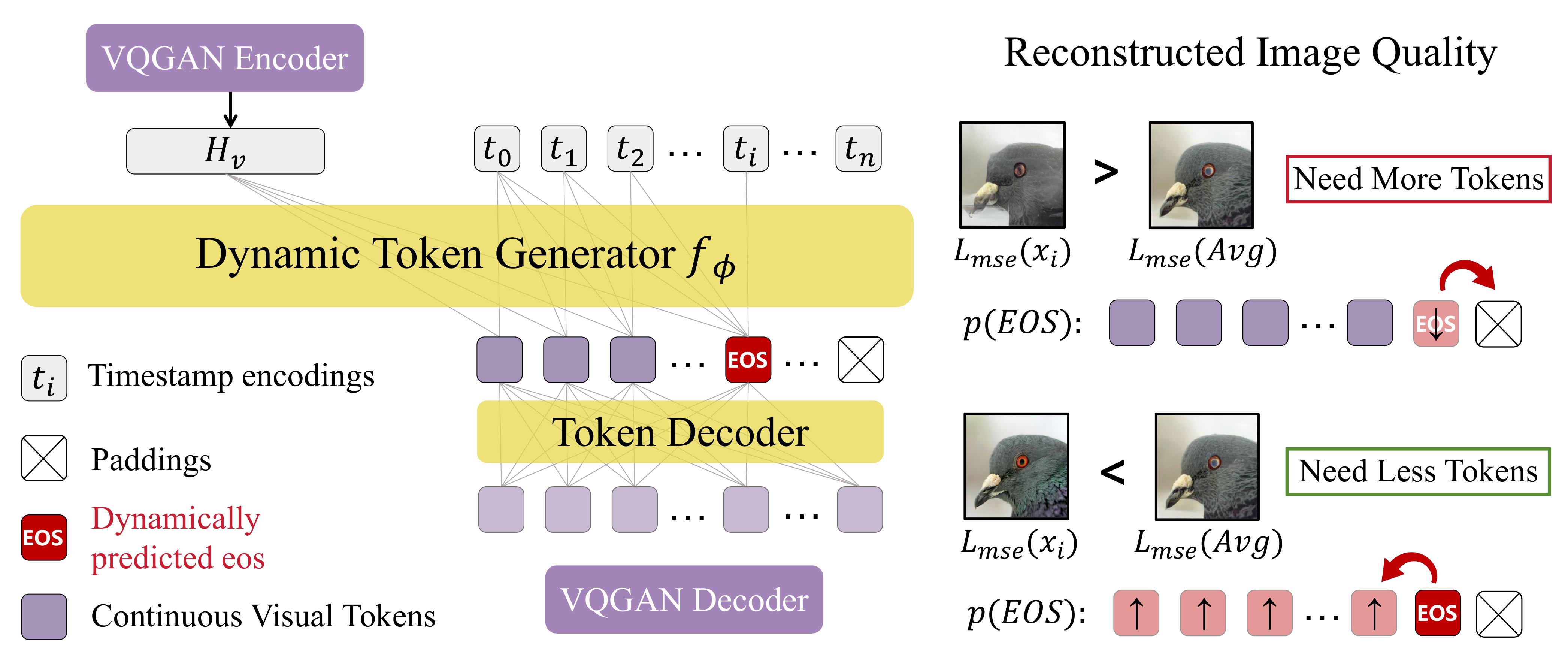}
    \caption{\textbf{Dynamic Tokenizer.} }
    \label{fig:DOVE_Framework}
    \vspace{-12pt}
\end{figure}

An overview of our model is shown in Figure~\ref{fig:DOVE_Framework}. Our model consists of four main components: VQGAN Encoder, VQGAN Decoder, transformer-based dynamic token generator, and transformer-based token decoder. We use 70M transformer~\cite{biderman2023pythia} as the backbone for both the autoregressive token generator and a non-autoregressive version for token decoder. 

For each image \( X_v \), the VQGAN Encoder converts the visual information into a fixed-length token sequence \( H_v \). Timestamp encodings \( t_1, t_2, \dots, t_n \), generated using periodic embeddings such as sinusoidal encodings~\cite{vaswani_attn}, are then appended to \( H_v \). This combined sequence is input into the dynamic token generator \( f_{\phi} \). To enable sequential token generation, we restrict each position to attend only to its current or preceding timestamps. The dynamic token generation process from timestamp \( t_0 \) to \( t_i \) is defined as:

\vspace{-8pt}
\begin{equation}
D = f_{\phi}(H_v, t_1, t_2, \dots, t_i) = (d_1, d_2, \dots, d_i)
\end{equation}
\vspace{-10pt}

where \( D \) denotes the generated token sequence, and \( d_i \) is the token produced by the model at \( t_i \). We introduce dynamic length variation by detecting the EOS token from the model’s discrete output and replacing all visual token (continuous latent outputs) from that position onward with zero vectors. Since the EOS token can appear at any position, the length of the generated token sequence can vary based on the complexity of the image. We use an additional non-autoregressive token decoder \( g_{\phi} \) to decode the padded dynamic vision token sequence and feed it to the final VQGAN decoder.




\subsection{Dynamic Image Reconstruction}

\begin{wraptable}{r}{0.5\textwidth}
\vspace{-16pt}
\small

\rule{\linewidth}{0.8pt}
\begin{algorithmic}
\Statex \textbf{Define:} Image $X_v$, max tokens $K$, window $W$, weights $\lambda_{\text{rec}}, \lambda_{\text{eos}}$, time encodings $T$
\vspace{5px}
\State $H_v \gets \text{VQGAN\_Encoder}(X)$
\State Initialize $\mathrm{EMA}_{\text{rec}}\gets 0$
\vspace{2px}
\For{each training iteration}
  \State $D\gets[\,\,]$, $i\gets1$
  \vspace{2px}
  \While{$i \le K$}
    \State $d_i \gets f_\phi(H_v, T_{1:i})$ \,\small{\textit{(generating token)}}
    \State append $d_i$ to $D$, $i\gets i+1$
  \EndWhile
  \vspace{2px}
  \State Find the first index $j$ such that $D[j] = \text{EOS}$
  \vspace{2px}
  \If{such $j$ exists}
    \For{$k = j+1$ to $K$}
      \State $D[k] \gets 0$
    \EndFor
  \EndIf
  \vspace{2px}
  \State $\hat X\gets \text{VQGAN\_Decoder}\bigl(g_\phi(D)\bigr)$
  \State Compute $L_{\text{rec}}$ via Eq.~\eqref{eq:rec_loss}
  \State Update $\mathrm{EMA}_{\text{rec}}$ over the last $W$ losses
  \vspace{2px}
  \If{$L_{\text{rec}} > \mathrm{EMA}_{\text{rec}}$}
    \State $L_{\text{eos}}\gets p_{\text{eos}}(i)$
  \Else
    \State $L_{\text{eos}}\gets -\frac{1}{i-1}\sum_{j=1}^{i-1}p_{\text{eos}}(j)$
  \EndIf
  \vspace{2px}
  \State $L_{\text{total}}\gets \lambda_{\text{rec}}\,L_{\text{rec}} + \lambda_{\text{eos}}\,L_{\text{eos}}$
  \State Update parameters $\phi$ using $\nabla_\phi L_{\text{total}}$
  \vspace{2px}
\EndFor
\end{algorithmic}
\vspace{-3pt} 
\rule{\linewidth}{1.0pt}
\vspace{-10pt} 
\caption{Training Pseudocode}\label{alg:DOVE}
\vspace{6.8pt}
\end{wraptable}
A more complex image, which contains richer and finer-grained details, will require more tokens to capture all its visual information compared to a simpler one. By learning when to generate EOS, the model can adaptively produce a token sequence that is just long enough to capture the image's essential visual content.

We jointly train all components of the model. Following the training strategy of VQGAN~\cite{esser2021taming}, we adopt a combination of mean squared error (MSE) loss and perceptual loss to supervise the image reconstruction process. A lightly weighted adversarial (GAN) loss is also applied to enhance the realism of reconstructed images. The final reconstruction loss \( L_{\text{rec}} \) between the input image \( X_v \) and the reconstructed image \( \hat{X}_v \) is defined as:

\vspace{-8pt}
\begin{equation}\label{eq:rec_loss}
L_{\text{rec}} = \lambda_{\text{mse}} \cdot L_{\text{mse}} + \lambda_{\text{perc}} \cdot L_{\text{perc}} + \lambda_{\text{gan}} \cdot L_{\text{gan}}
\end{equation}
\vspace{-20pt}

During training, we set the weighting factors to \(\lambda_{\text{mse}} = 1\), \(\lambda_{\text{perc}} = 0.1\), and \(\lambda_{\text{gan}} = 5 \times 10^{-10}\) to prevent hallucination. In parallel with improving reconstruction quality, we guide the model to adaptively adjust the length of the generated token sequence through EOS prediction. Specifically, we use the average reconstruction loss \(L_{\text{rec}}\) over the previous 100 training steps as a dynamic threshold. For a given sample, if its current reconstruction loss is lower than the threshold, it indicates that fewer tokens are sufficient for satisfactory reconstruction, and we encourage earlier EOS prediction by maximizing the EOS probabilities at all preceding positions. Conversely, if the reconstruction loss exceeds the threshold, it suggests that more tokens are needed, and we minimize the EOS probability at the current position.

We denote the predicted EOS probability at position \( i \) as \( p_{\text{eos}}(i) \), where \( m \) indicates the current EOS position.  The token length control loss is defined as:

\vspace{-8pt}
\begin{equation}\label{eq:eos_loss}
L_{\text{eos}} =
\begin{cases}
p_{\text{eos}}(m), & \text{if } L_{\text{rec}} > \text{Threshold} \\
-\dfrac{1}{m-1} \sum\limits_{i=1}^{m-1} p_{\text{eos}}(i), & \text{if } L_{\text{rec}} \leq \text{Threshold}
\end{cases}
\end{equation}
\vspace{-8pt}

Finally, we jointly optimize \(L_{\text{rec}}\) and \(L_{\text{eos}}\) to guide the model in dynamically reconstructing the image. The overall training loss is defined as:

\vspace{-8pt}
\begin{equation}
L_{\text{total}} = \lambda_{\text{rec}} L_{\text{rec}} + \lambda_{\text{eos}} L_{\text{eos}}
\end{equation}
\vspace{-4pt}

where \(\lambda_{\text{rec}}\) and \(\lambda_{\text{eos}}\) are the corresponding weighting coefficients. To facilitate faster convergence, we initially set \(\lambda_{\text{eos}}\) to a small value and gradually increase it during training, allowing the model to first focus on accurate reconstruction before learning to adaptively control the token sequence length.








\vspace{-5pt}
\subsection{Q-DOVE: Query-conditioned Tokenization}

\begin{figure}[!h]
    \centering
    \includegraphics[width=\linewidth]{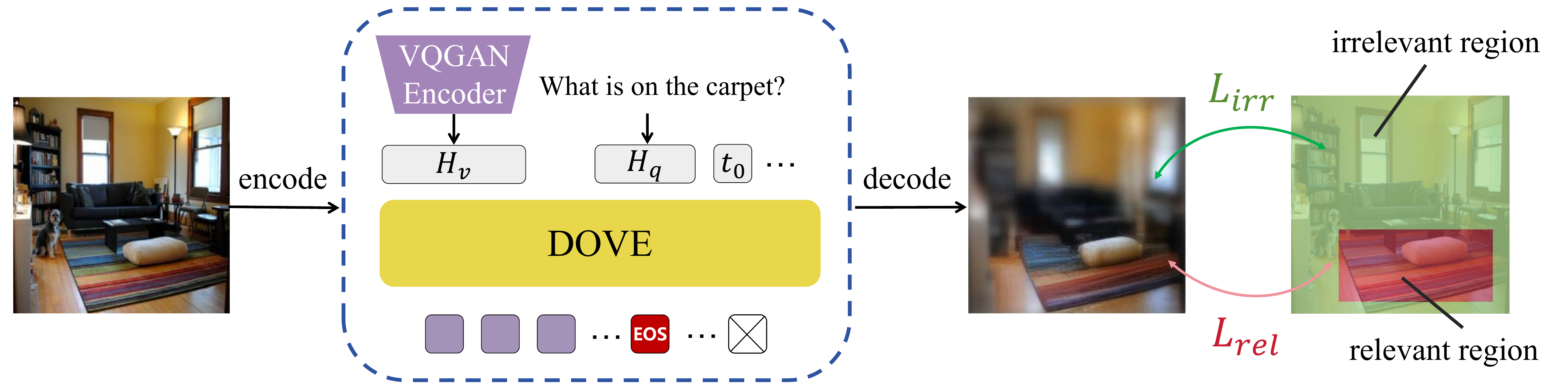}
    \caption{\textbf{Query Conditioning.} DOVE is trained with a bounding-box based loss, learning to focus its dynamic token resources on representing query-relevant image regions.}
    \label{fig:Query_Condition}
    \vspace{-7pt}
\end{figure}

We extend DOVE to Q-DOVE for use in text-conditioned vision and language domains (Figure \ref{fig:Query_Condition}), allowing it to dynamically adapt image representations in a query-dependent manner. Q-DOVE is trained to focus image representation resources on image regions relevant to a given query. 

Given a supervised dataset of images paired with text queries and bounding boxes encapsulating their answers, we modify the reconstruction loss to focus over image regions within each example's set of bounding boxes $S_{bb}$.
Specifically, we upsample each image region contained by a bounding box $b^i\in S_{bb}$ to an image $I_{bb}^i$ and compute the reconstruction loss over it as in Eq. \ref{eq:rec_loss}:
\begin{equation}
    L_\text{rel}^i = L_\text{rec}(I_{bb}^i) 
\end{equation}
In order to encourage the model to maintain some fidelity over the region outside of the bounding boxes, we also compute the MSE loss over $I_o$, the complement of $S_{bb}$: 
\begin{equation}
    L_\text{irr} = L_\text{mse}(I_o)
\end{equation}
The final loss averages over relevant regions and weighs loss over the irrelevant region down by $\lambda_o$:
\begin{equation}
    L_\text{qry} = \frac{\sum_{b^i\in S_{bb}}L_\text{rel}^i}{|S_{bb}|} + \lambda_o \cdot L_\text{irr}
\end{equation}
In our experiments, we set $\lambda_o$ to 1e-10.  
To compute $L_\text{eos}$, we employ the same procedure as in Eq. \ref{eq:eos_loss}, comparing $L_{rel}$ to a threshold determined by its average loss over previous training steps. If $L_{irr}$  falls below the threshold, we introduce an additional penalty $L_\text{pen}$ to explicitly encourage the model to generate the EOS token earlier:
$
    L_\text{pen} = -\dfrac{1}{m-1} \sum\limits_{i=1}^{m-1} p_{\text{eos}}(i)
$.

Our supervised masking strategy yields a dual benefit, allowing the model to learn both where to look and how much information to encode from image regions relevant to inputted queries. 
Bounding boxes are only used during training. 

%% file: sec/4_exp.tex
\section{Experiments}

In this section,  We evaluate our approach at multiple levels, including the quality of the generated vision tokens (e.g., image reconstruction and token length distribution), as well as their effectiveness in downstream vision-language tasks. The results demonstrate that our model achieves high reconstruction quality with significantly fewer tokens, while capturing richer semantic information compared to static autoencoder-based tokenization methods. We further investigate the phenomenon of emergent semantics in Section 3.4.

\subsection{Experimental Setup}

\textbf{Training Details.} We use a pretrained VQGAN~\cite{esser2021taming} with a codebook size of 8192 and a lightweight Pythia-70M~\cite{biderman2023pythia} language model as the backbone of our framework. The model is fine-tuned on ImageNet-1K~\cite{cui2023scaling} for 20 epochs using two NVIDIA RTX 4090 GPUs. For the query-conditioned variant, we conduct an additional 5 epochs of training on the Visual Genome~\cite{krishna2017visual} and Open Images~\cite{kuznetsova2020open} datasets. We directly use the provided questions and region-level captions in Visual Genome as textual queries to guide the model in reconstructing content within specified bounding boxes, while ignoring irrelevant regions. Since Open Images does not offer region-level descriptions or questions, we instead construct text queries from relation graph annotations—for example, ``a cup on a table''—and define the target region by concatenating the bounding boxes of the associated objects. To improve the model's generalization ability, we randomly replace 50\% of the training text queries with the string ``null'', and train the model to reconstruct the entire image when this placeholder is provided as input.

\textbf{Baselines.} We compare our model against several state-of-the-art encoder-decoder frameworks, including TiTok\cite{yu2024image} and VQGAN. We choose VQGAN with an output length of 256 tokens. For TiTok, we consider three variants with token lengths of 32, 64, and 128. We also include ALIT~\cite{duggal2024adaptive}, a dynamic vision encoder trained via recurrent distillation from VQGAN. Unlike our method, however, ALIT only supports token lengths that are multiples of a fixed stride (e.g., 32). All models are trained on ImageNet-1K under the same configuration to ensure a fair comparison.

\subsection{Token-Level Evaluation}

\begin{figure}[!h]
    \centering
    \includegraphics[width=\linewidth]{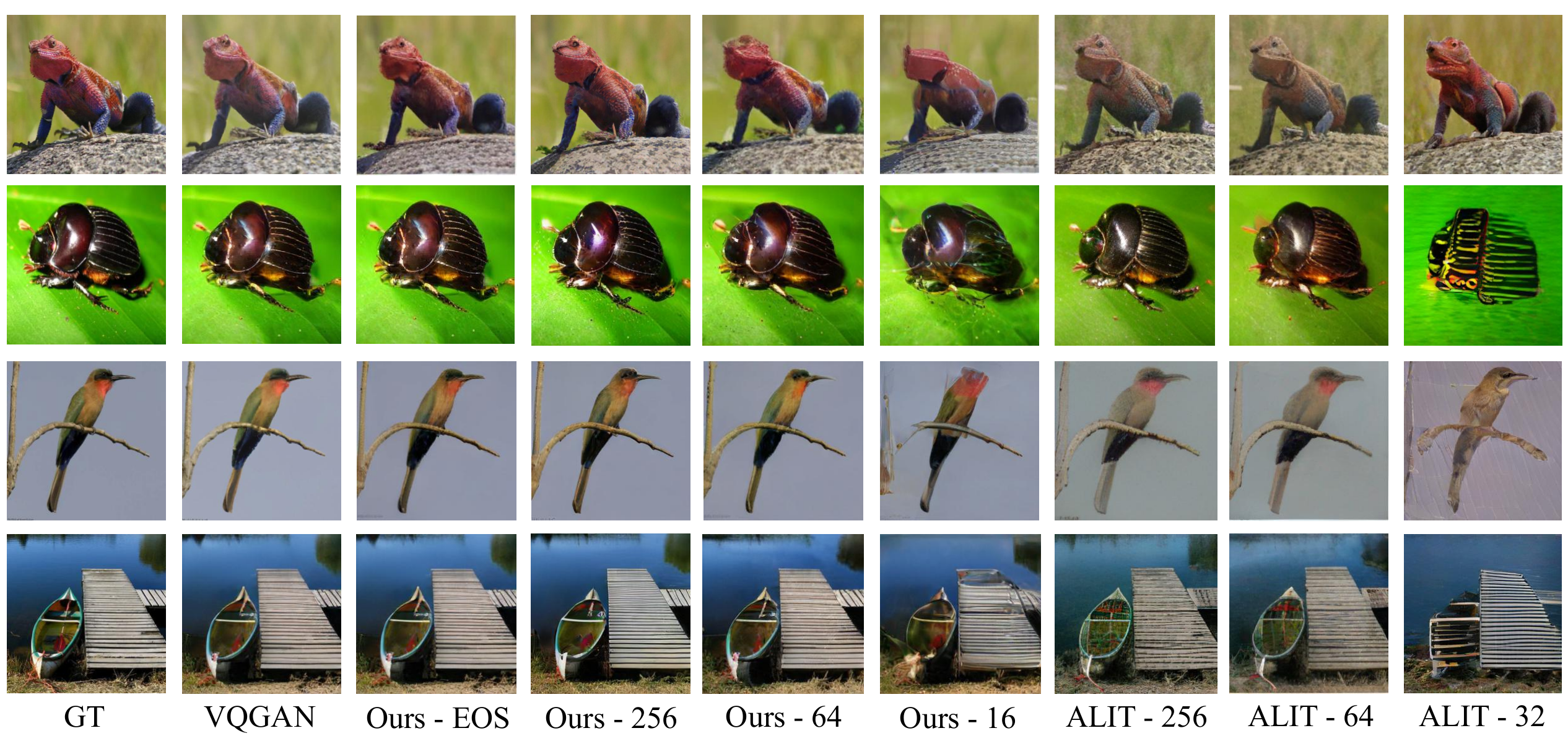}
    \caption{Reconstructed images on ImageNet-1K using different methods. As the token length increases, our method produces progressively clearer reconstructions with more visual details.}
    \label{fig:DVT_Case}
    \vspace{-12pt}
\end{figure}

\textbf{Image Reconstruction Quality.} We report FID scores of the reconstructed images across varying token lengths. Our results show that as the token length increases, the reconstruction quality of our model consistently improves. At all evaluated token lengths, our method outperforms ALIT. This advantage becomes especially clear at lower token counts. ALIT often generates hallucinated content, including severe object distortions. For example, when the token length is limited to 32, the reconstructed chameleon and beetle exhibit noticeable deformations (Figure~\ref{fig:DVT_Case}). In contrast, our model produces slightly blurry but structurally and semantically faithful reconstructions. When using the full token length of 256, our method surpasses VQGAN on the COCO and WIT datasets. Detailed results are provided in Table~\ref{tab:reconstruction_results}.

\begin{table*}[ht]
\centering
\resizebox{\textwidth}{!}{%
\begin{tabular}{lcccccccccccccc}
\toprule
\multirow{2}{*}{\textbf{Approach}} 
& \multicolumn{8}{c}{\textbf{ImageNet100}} 
& \multicolumn{3}{c}{\textbf{COCO}} 
& \multicolumn{3}{c}{\textbf{Wikipedia (WIT)}} \\
\cmidrule(lr){2-9} \cmidrule(lr){10-12} \cmidrule(lr){13-15}
 & 32 & 64 & 96 & 128 & 160 & 192 & 224 & 256 
 & 32$^\#$ / 64 & 128 & 256 
 & 32$^\#$ / 64 & 128 & 256 \\
\midrule
TiTok-L-32      & 11.60 & -    & -    & -    & -    & -    & -    & -    & 14.18$^\#$ & -     & -     & 53.57$^\#$ & -     & -     \\
TiTok-B-64      & -     & 8.22 & -    & -    & -    & -    & -    & -    & 9.15       & -     & -     & 42.86      & -     & -     \\
TiTok-S-128     & -     & -    & - & 8.22    & -    & -    & -    & -    & -          & 9.15     & -     & -          & 38.16 & -     \\
VQGAN          & -     & -    & -    & -    & -    & -    & -    & 7.04    & -       & -     & 7.77     & -          & -     & 31.27  \\
ALIT          & 22.31 & 15.92 & 13.08 & 11.45 & 10.01 & 9.12 & 8.37 & 8.06 & 22.01 & 13.98 & 9.51 & 61.32 & 47.52 & 38.10 \\
DOVE  & 18.91 & 11.46 & 10.84 & 9.28 &  8.61 & 8.25 & 7.96 & 7.73 & 15.50 & 9.83 & 7.54 & 14.83 & 8.56 & 7.84      \\
\bottomrule
\end{tabular}%
}
\caption{FID scores (↓) across the ImageNet100, COCO, and WIT datasets. Our method consistently outperforms ALIT across all token lengths, and achieves comparable or even better results than VQGAN and TiTok at several lengths.}

\label{tab:reconstruction_results}
\vspace{-7pt}
\end{table*}

\textbf{Classification.} We evaluate the representation quality of DOVE as an off-the-shelf, frozen backbone across three standard recognition benchmarks, including CIFAR-100~\cite{krizhevsky2009learning}, ImageNet-100~\cite{deng2009imagenet}, and STL-10~\cite{N/A_2024}. Specifically, we train a lightweight MLP classifier on top of the frozen features, using both mean and max pooling over the final layer representations. As the number of tokens increases, the classification accuracy of both DOVE and ALIT steadily improves.
Our approach consistently outperforms all other vision tokenizers by a substantial margin. Even when using as few as 32 tokens, it achieves higher classification accuracy than all competing methods. We attribute this advantage to our dynamic reconstruction training objective, which enables the model to capture additional semantic information during representation learning. This is further evidenced by the linear probing and PCA-based zero-shot segmentation results presented in Section 3.4.
\vspace{-3pt}

\begin{figure}[!h]
    \centering
        \begin{subfigure}{0.31\linewidth}
        \centering
        \includegraphics[width=\linewidth]{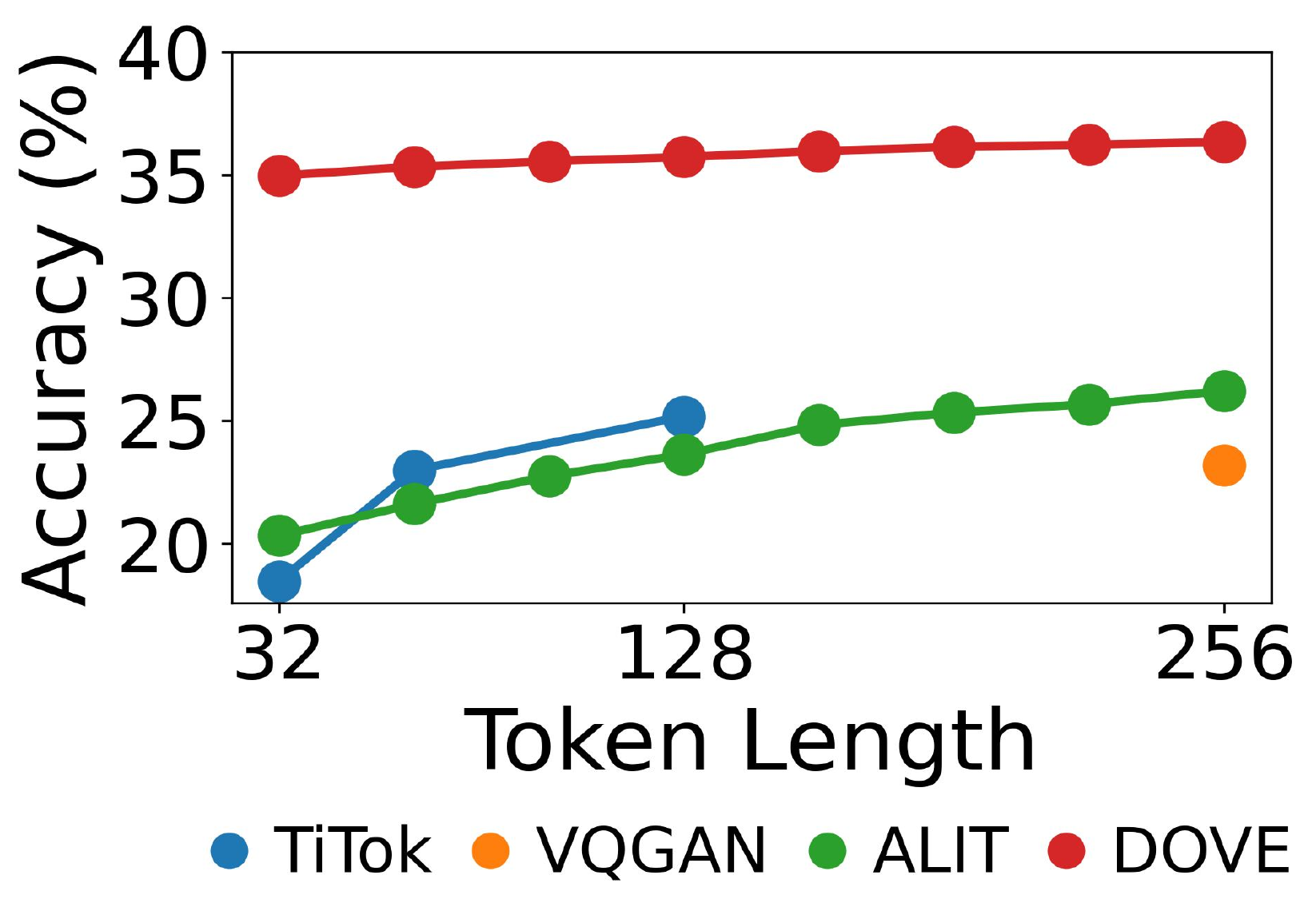}
        \caption{CIFAR100}
        \label{fig:Length_distribution}
    \end{subfigure}
    \hfill
    \begin{subfigure}{0.31\linewidth}
        \centering
        \includegraphics[width=\linewidth]{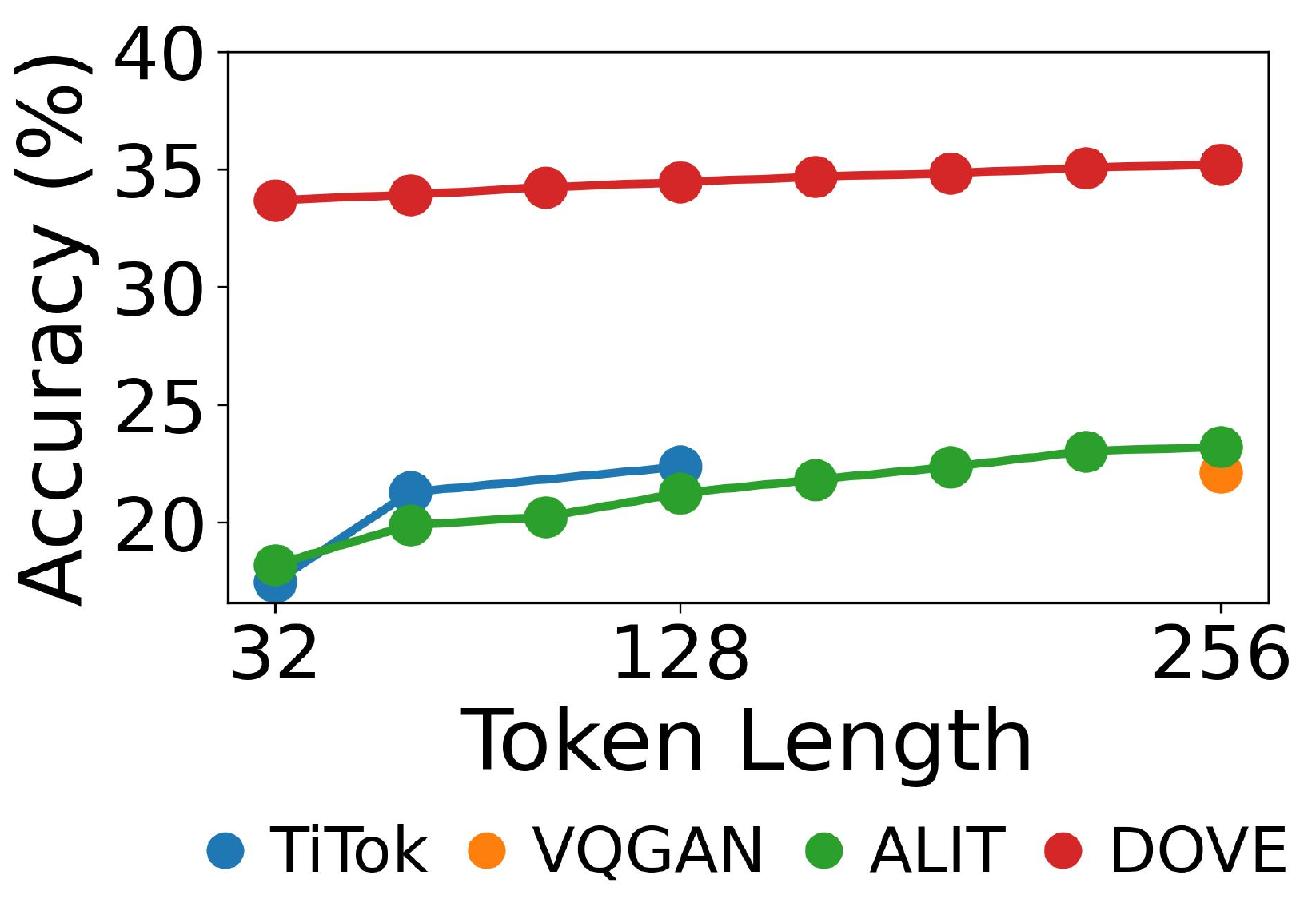}
        \caption{ImageNet100}
        \label{fig:Token_Length_Reconstruction_Loss}
    \end{subfigure}
    \hfill
    \begin{subfigure}{0.31\linewidth}
        \centering
        \includegraphics[width=\linewidth]{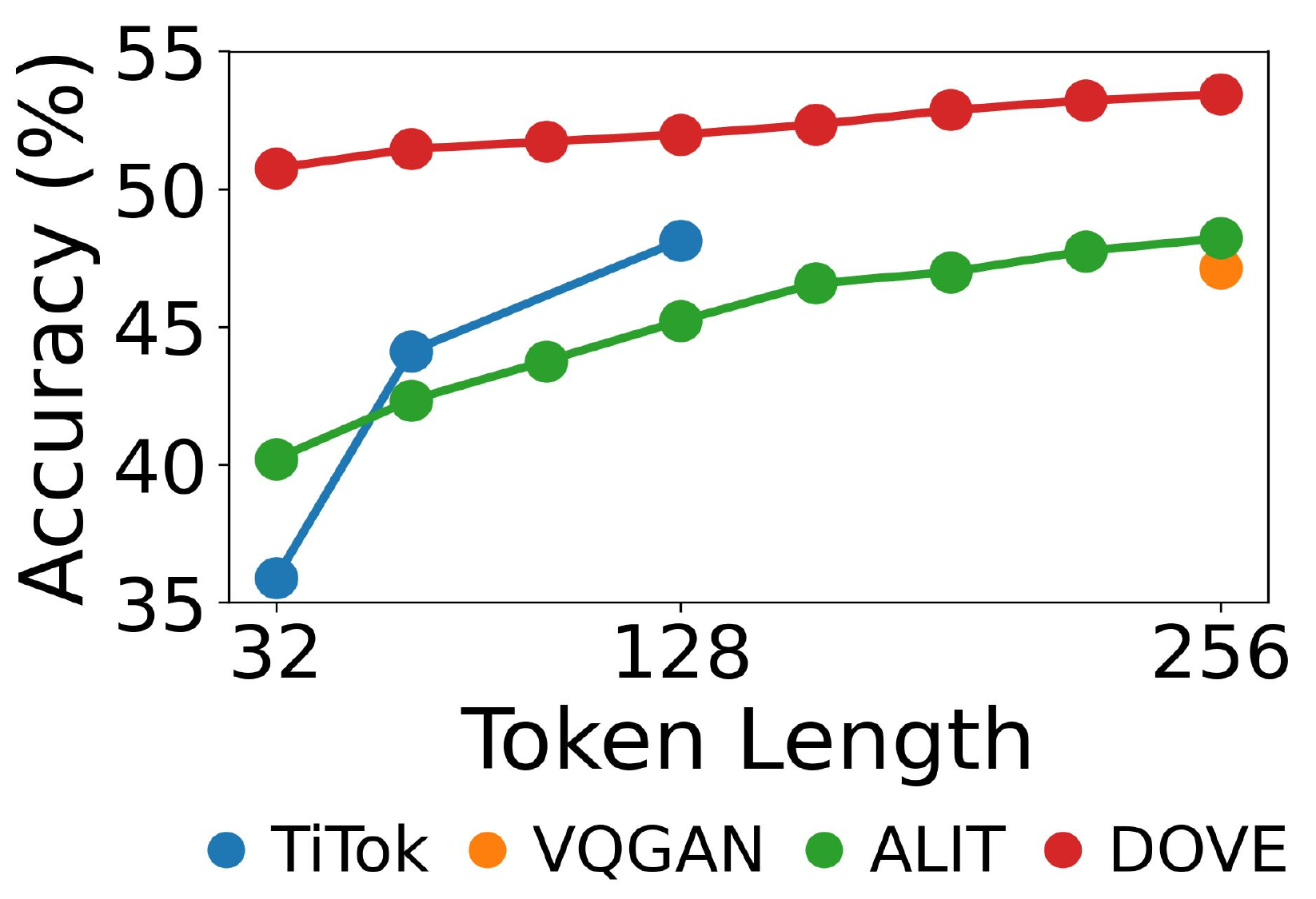}
        \caption{STL-10}
        \label{fig:Complexity}
    \end{subfigure}
    \caption{Classification accuracy with different visual tokenizers under varying token lengths. DOVE consistently outperforms all baselines across all lengths.}
    \vspace{-6pt}
    \label{fig:Linear_Prob_Classfication}
\end{figure}
\vspace{-2pt}

\textbf{Token Length Distribution.} Unlike ALIT, our model explicitly supports a mechanism for generating arbitrary-length token sequences at inference time. We analyze the distribution of token sequence lengths (i.e., EOS positions) generated by DOVE. As shown in Figure~\ref{fig:Length_distribution}, most sequences are shorter than 100 tokens, with smaller peaks around 150 and 250. We randomly sample 5,000 images from the MS COCO 2017 validation set~\cite{lin2014microsoft} and compute the reconstruction loss across different token lengths.
Figure~\ref{fig:Token_Length_Reconstruction_Loss} shows that reconstruction loss decreases as token length increases. This decline is steepest between 0 and 100 tokens, and becomes more gradual beyond that.
To further investigate the relationship between token length and image content, we calculate the complexity of input images using Laplacian variance~\cite{bansal2016blur} and analyze the correlation between image complexity and the length of the generated token sequences. As shown in Figure~\ref{fig:Complexity}, by encouraging samples with lower reconstruction quality to delay the EOS position and those with higher quality to emit EOS earlier during training, DOVE naturally learns to allocate longer token sequences to more complex images, while assigning shorter sequences to simpler ones. The Pearson correlation coefficient between image complexity and token sequence length is 0.742.
\vspace{-5pt}
\begin{figure}[!h]
    \centering
        \begin{subfigure}{0.32\linewidth}
        \centering
        \includegraphics[width=\linewidth]{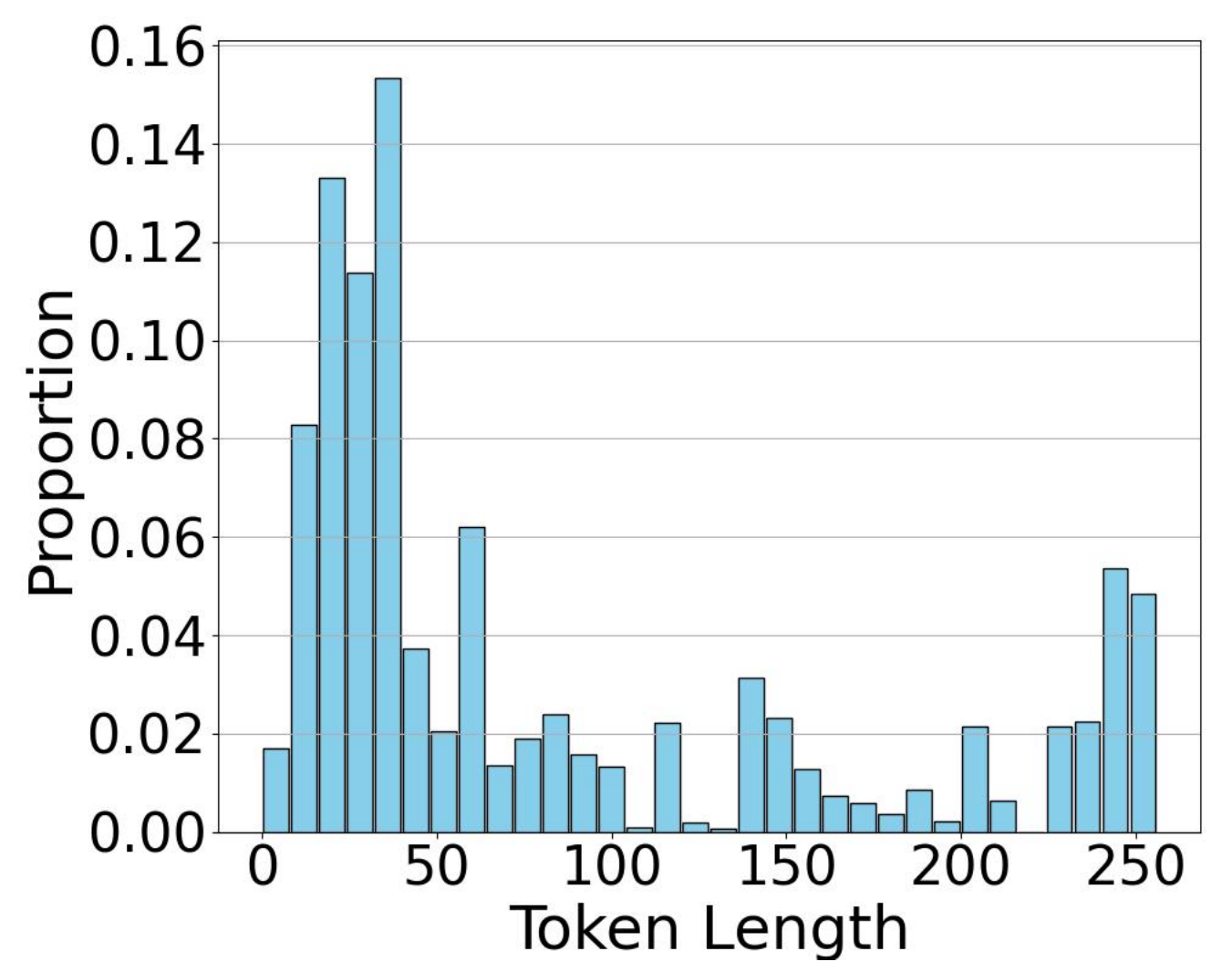}
        \caption{Distribution of token sequence lengths (i.e.,EOS positions) generated by DOVE.}
        \label{fig:Length_distribution}
    \end{subfigure}
    \hfill
    \begin{subfigure}{0.28\linewidth}
        \centering
        \includegraphics[width=\linewidth]{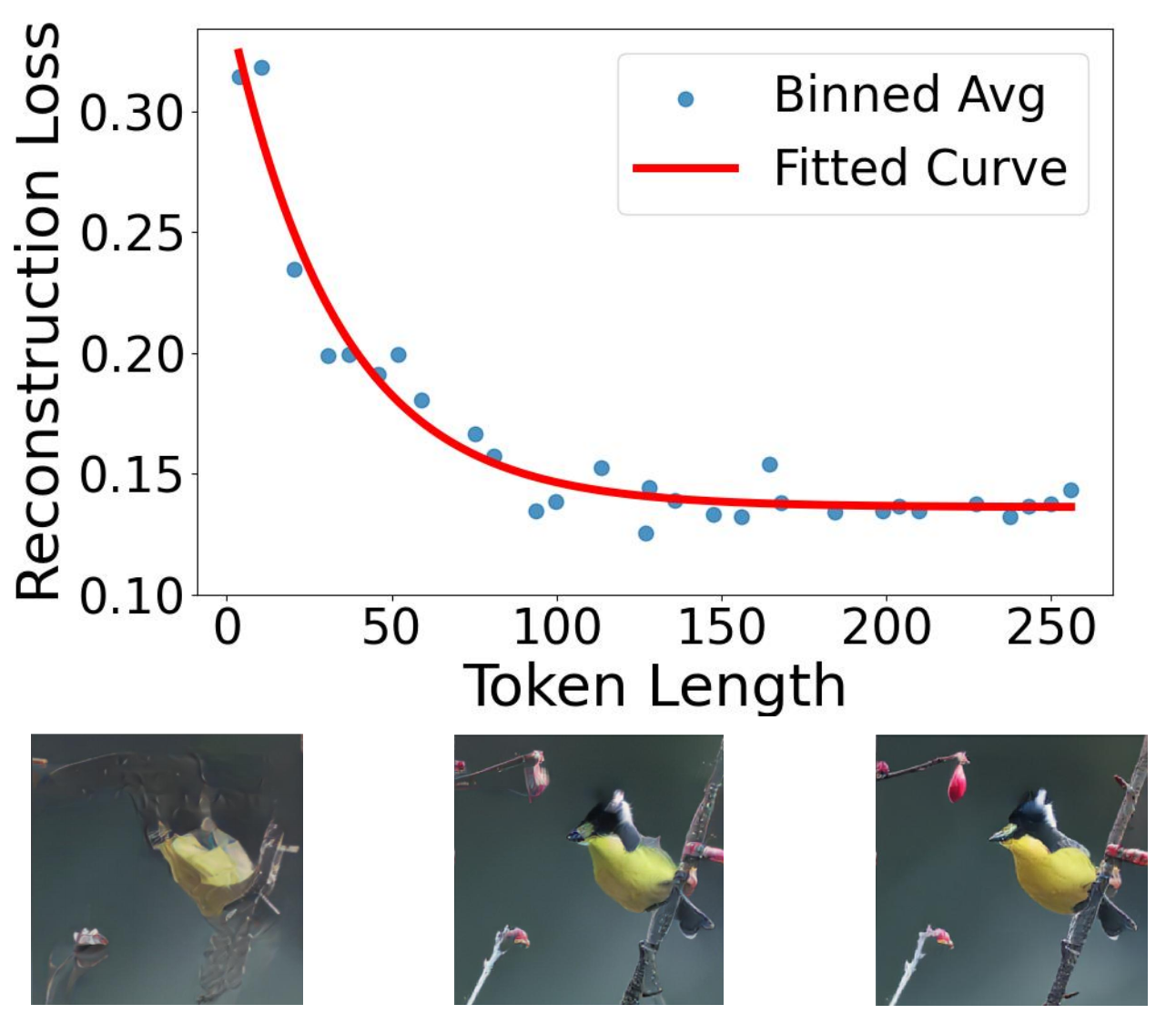}
        \caption{The relation between token length and reconstruction loss across different input samples.}
        \label{fig:Token_Length_Reconstruction_Loss}
    \end{subfigure}
    \hfill
    \begin{subfigure}{0.28\linewidth}
        \centering
        \includegraphics[width=\linewidth]{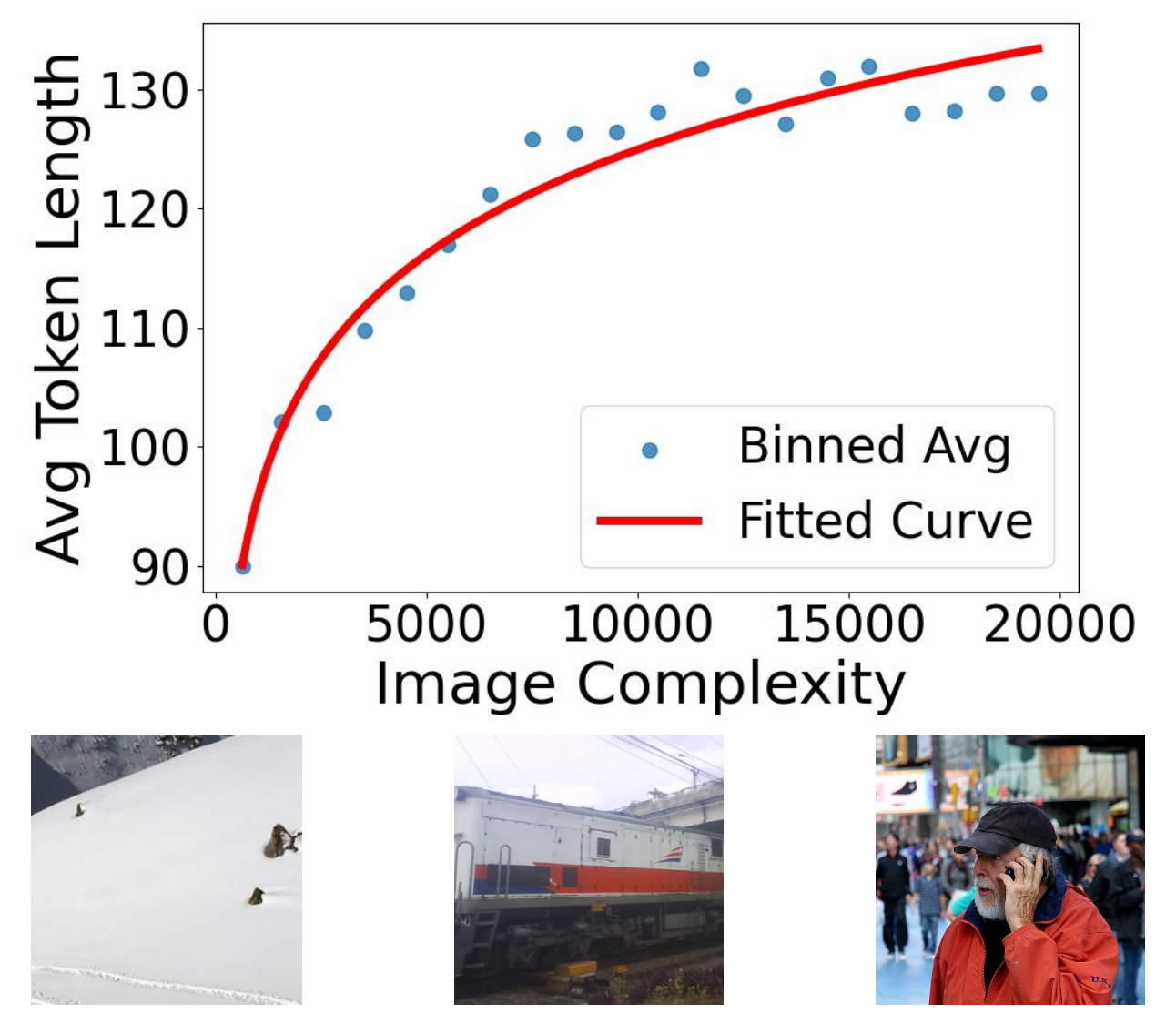}
        \caption{The relation between token sequence lengths (i.e.,EOS positions) and image complexity.}
        \label{fig:Complexity}
    \end{subfigure}
    \caption{Token length analysis}
    \vspace{-10pt}
    \label{fig:length}
\end{figure}

\subsection{Downstream Vision-Language Task Evaluation}
\begin{figure}[!h]
    \centering
    \includegraphics[width=\linewidth]{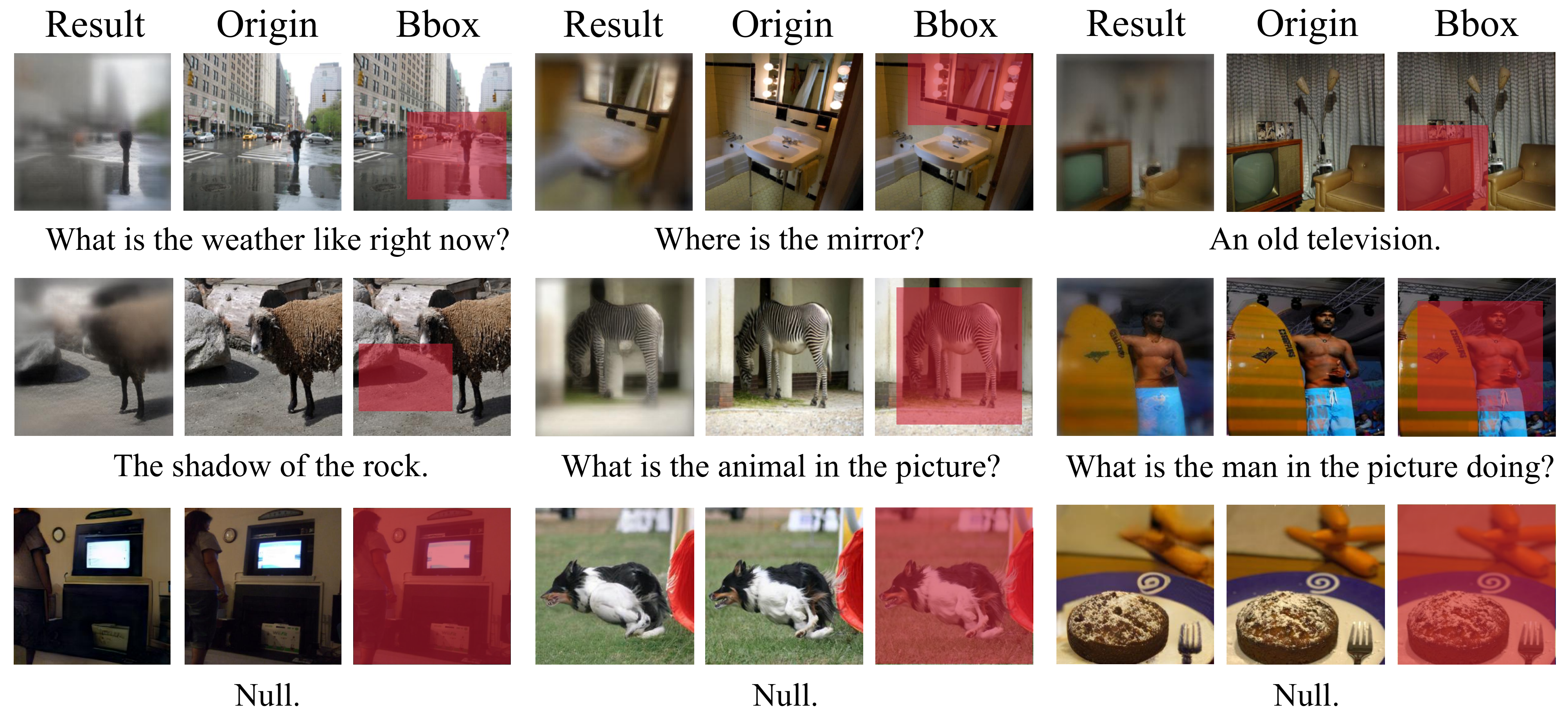}
    \caption{
    Reconstructed images from the Q-DOVE. When the text query is set to ``null'', the model reconstructs the entire image. When a query is provided, the model focuses on query-relevant regions.
    }

    \label{fig:query_conditioned}
    \vspace{-7pt}
\end{figure}

\vspace{-2pt}
\textbf{Query-conditioned Tokenization.} We visualize the behavior of our query-conditioned DOVE (Q-DOVE) on the Visual Genome dataset. Figure~\ref{fig:query_conditioned} presents several examples. The results show that when the input query is ``null'', the model clearly reconstructs the entire image. In contrast, when a relevant question or description is provided, the reconstruction focuses on the semantically related regions and produces lower frequency outputs for background. This task-driven compression even further reduces the average token sequence length. We then evaluate Q-DOVE and the original DOVE model as vision encoders in downstream vision-language tasks.

\vspace{-5pt}

\textbf{Visual Question Answering Evaluation.} To evaluate the quality of our model's token representations, we replace the vision encoder in a vision-language model with different visual representation methods and evaluate them on downstream vision-language tasks. We adopt Vicuna-7B-v1.5~\cite{liu2023llava} as the language model, interfacing it with a two-layer MLP that maps the vision encoder outputs to the language model input space. Following the training strategy of AIM V2~\cite{fini_multimodal-autoregressive}, we set the learning rate of the language model to 2e-5 and that of the adapter layers to 2e-4. This setup enables joint fine-tuning in a single-stage training process. We fine-tune the model with different vision encoders for one epoch on the 665K mixed VQA dataset used in LLaVA~\cite{liu2023llava}. The model is evaluated on a broad set of benchmarks, including VQAv2~\cite{goyal2017makingvvqamatter}, GQA~\cite{ainslie2023gqatraininggeneralizedmultiquery}, OK-VQA~\cite{marino2019okvqavisualquestionanswering}, TextVQA~\cite{singh2019vqamodelsread}, DocVQA~\cite{mathew2021docvqadatasetvqadocument}, InfoVQA~\cite{mathew2021infographicvqa}, ChartQA~\cite{masry2022chartqabenchmarkquestionanswering}, and ScienceQA~\cite{lu2022learnexplainmultimodalreasoning}. 

Results show that the VLM equipped with DOVE significantly outperforms other models across all datasets. Moreover, integrating Q-DOVE further improves the accuracy. By leveraging DOVE's EOS token as a truncation point, we achieve a substantial reduction in token count with performance comparable to the full set of 256 tokens. For Q-DOVE, we include two input strategies for the vision encoder: providing the actual question or directly inputting a ``null''. While the ``null'' setting yields slightly better performance than using the question—which filters out task-irrelevant regions—the question-guided strategy achieves comparable accuracy while further reducing the token length.

We also measure the inference time and floating-point operations (FLOPs) of each model, as shown in Table~\ref{tab:time}. Both our method and ALIT can effectively reduce FLOPs by shortening the length of the visual token sequence. However, due to ALIT's use of recurrent distillation, where dynamic tokens are generated through multiple passes over VQGAN tokens, its inference speed is adversely affected despite the reduced sequence length. In contrast, our method relies on a single forward pass, resulting in much faster inference.

\begin{table*}[!h]
\centering
\resizebox{\textwidth}{!}{%
\begin{tabular}{lccccccccccc}
\toprule
\textbf{Model} & \# Token Count & VQAv2 & GQA & OKVQA & TextVQA & DocVQA & InfoVQA & ChartQA & ScienceQA \\
\midrule
Titok & 128 (S) & 43.3 & 38.8 & 38.6 & 14.3 & 8.1 & 17.0 & 11.8 & 67.1 \\
\midrule
VQGAN & 256 & 40.2 & 38.1 & 37.7 & 14.3 & 8.2 & 16.3 & 11.1 & 66.3 \\
\midrule
\multirow{4}{*}{ALIT} & 32 & 38.4 & 37.6 & 35.6 & 14.2 & 7.8 & 16.0 & 11.4 & 66.0 \\
& 64 & 39.7 & 38.0 & 36.4 & 14.3 & 8.1 & 16.2 & 11.6 & 66.2 \\
& 128 & 41.0 & 38.0 & 37.2 & 14.3 & 8.2 & 16.3 & 11.7 & 66.5 \\
& 256 & 43.8 & 38.3 & 37.8 & 14.3 & 8.2 & 16.5 & 12.0 & 66.8 \\
\midrule
\multirow{5}{*}{DOVE} & 32 & 50.3 & 47.2 & 42.2 & 14.6 & 7.9 & 18.4 & 11.2 & 69.6 \\
& 64 & 51.8 & 50.2 & 43.5 & 14.9 & 8.2 & 18.8 & 12.1 & 71.7 \\
& 128 & 52.0 & 50.7 & 44.8 & 15.0 & 8.2 & 19.1 & 12.4 & 72.5 \\
& 256 & 52.4 & 51.8 & 46.2 & 15.0 & 8.4 & 19.4 & 12.6 & 72.8 \\
& 121.6 (Avg) & 52.2 & 51.4 & 46.0 & 15.0 & 8.2 & 19.2 & 12.6 & 72.6 \\
\midrule
\multirow{3}{*}{Q-DOVE} & 256\# & \textbf{55.0} & \textbf{53.2} & \textbf{46.7} & \textbf{15.3} & \textbf{8.6} & \textbf{19.7} & \textbf{12.8} & \textbf{74.8} \\
& 256 & 53.9 & 52.6 & 46.2 & 15.2 & 8.2 & 19.4 & 12.5 & 74.0 \\
& 82.4 (Avg) & 52.8 & 52.1 & 46.0 & 15.2 & 8.2 & 19.2 & 12.4 & 73.1 \\
\bottomrule
\end{tabular}
}
\caption{Performance comparison of VLMs equipped with different vision encoders. DOVE/Q-DOVE consistently achieves the best performance on most tasks. For Q-DOVE, ``\#'' indicates that the input query is set to ``null''; otherwise, the original question is used.}
\label{tab:time}
\vspace{-5pt}
\end{table*}

\begin{table}[h]
\centering
\resizebox{\textwidth}{!}{
\begin{tabular}{lccccccccc}
\toprule
\textbf{Model} & VQGAN-256 & ALIT-256 & ALIT-128 & ALIT-64 & ALIT-32 & DOVE-256 & DOVE-128 & DOVE-64 & DOVE-32 \\
\midrule
Speed ($\uparrow$) & $1.00 \times$ & $0.63\times$ & $0.82 \times$ & $0.88 \times$ & $0.92 \times$ & $0.96 \times$ & $1.14 \times$ & $1.19 \times$ & $1.26 \times$ \\
FLOPs (T, $\downarrow$) & 2.62 & 2.73 & 1.74 & 1.31 & 0.98 & 2.66 & 1.70 & 1.29 & 0.96 \\
\bottomrule
\end{tabular}
}
\vspace{1mm}
\caption{Inference speed and FLOPs (in teraflops) of different models. Inference speed is reported as the ratio relative to VQGAN, based on actual inference time measured on the VQAv2 test set.}
\label{tab:speed_memory}
\vspace{-12pt}
\end{table}

\subsection{Probing Emerging Semantics}

\begin{figure}[!h]
    \centering
    \includegraphics[width=\linewidth]{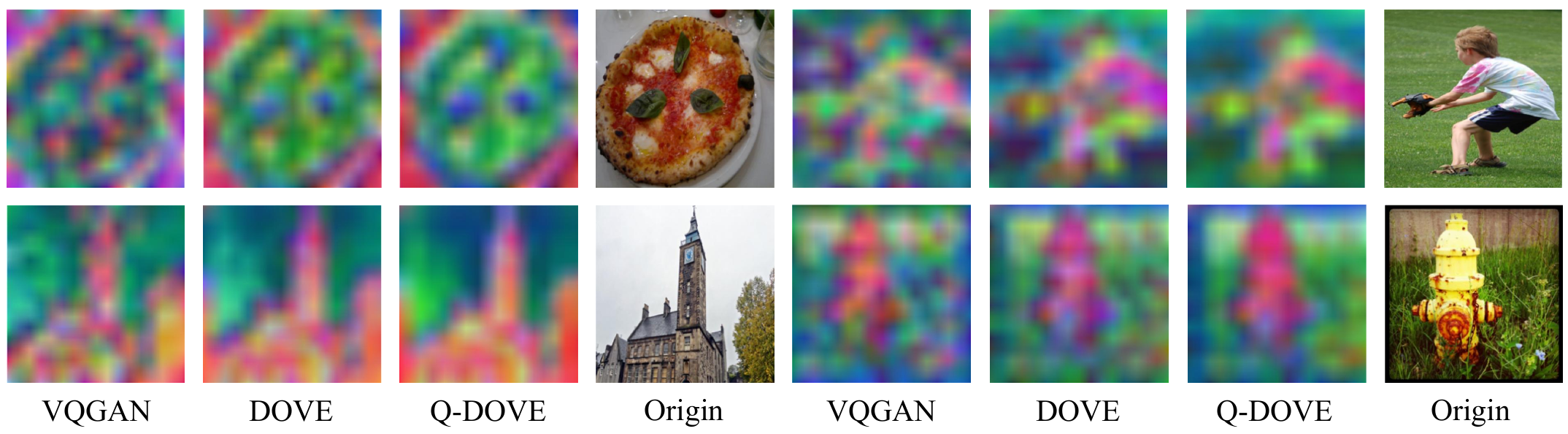}
    \caption{Semantics Visualization with PCA on latent features.}
    \label{fig:semantics}
    \vspace{-12pt}
\end{figure}

From previous experiments, we observe that the visual representations generated by DOVE significantly outperform those produced by fixed-length, autoencoder-based tokenization methods in both classification and downstream multimodal tasks. In this section, we further investigate this emergent semantic property through a series of analyses. Specifically, we evaluate the quality of the learned representations via linear probing on model's hidden layers instead of generated visual tokens and PCA-based image segmentation. We compare DOVE, Q-DOVE, and other fixed-length autoencoder-based tokenizers by conducting linear probing on seven benchmark datasets: CIFAR-10~\cite{krizhevsky2009learning}, CIFAR-100~\cite{krizhevsky2009learning}, DTD~\cite{cimpoi14describing}, FGVC~\cite{maji2013finegrainedvisualclassificationaircraft}, Food101~\cite{bossard14}, STL-10~\cite{coates2011analysis}, and SUN397~\cite{5539970}. For Q-DOVE, we set all text queries to “null” to simulate the unconditional setting. Table~\ref{tab:model-perf} shows that DOVE consistently outperforms other methods by a large margin across all datasets, and Q-DOVE further improves upon DOVE’s performance. To gain deeper insight into the structure of the learned representations, we apply PCA for dimensionality reduction and visualize the results in image space. As shown in Figure~\ref{fig:semantics}, DOVE yields more semantically coherent segmentations compared to VQGAN, while Q-DOVE exhibits even stronger semantic alignment and clarity.

\begin{table}[ht]
\centering
\resizebox{0.75\textwidth}{!}{%
\begin{tabular}{lcccccccc}
\toprule
Method          & CIFAR-10 & CIFAR-100 & DTD   & FGVC  & Food101 & STL-10 & SUN397 \\
\midrule
TiTok-32       & 24.87 &    6.11  &     9.46  &  1.95 &  3.81 &   23.23 &   4.44         \\
TiTok-64      & 25.95 &    7.34  &    10.74  &  2.61 &  4.53 &   28.06 &   5.23         \\
TiTok-128      & 18.33 &    3.10  &     6.80  &  2.34 &  3.05 &   20.25 &   3.02         \\
ALIT         & 41.08 &  16.87   &    26.96  &  4.47 & 14.47  &   42.15 &  20.94        \\
VQGAN         & 41.23 &   19.37  &    24.47  &  4.38 & 13.28 &   40.46 &  15.20         \\
\midrule 
DOVE           & 54.31 &   31.13  &    26.70  &  5.85 & 21.18 &   48.38 &  30.62         \\
Q-DOVE   & \textbf{56.44} &   \textbf{33.70}  &   \textbf{30.48}  &  \textbf{6.03} & \textbf{25.32} &   \textbf{54.86} &  \textbf{38.18}        \\
\bottomrule
\end{tabular}}
\vspace{2mm}
\caption{Linear probing performance (\%) of various models across benchmark datasets.}
\label{tab:model-perf}
\vspace{-20pt}
\end{table}


%% file: sec/2_related_works.tex
\section{Related Works}
\vspace{-5pt}
\textbf{Image Tokenization.}
Image tokenization methods represent images as discrete sets of patch embeddings. 
In ViT formulations~\cite{dosovitskiy2021an}, patch representations allow for efficient feature extraction with a transformer~\cite{vaswani_attn} in addition to direct compatibility with tokenized representations in other modALITies, such as text, through the use of projection layers~\cite{radford2021learning,liu2023llava}. 
Through vector quantization~\cite{van2017neural,razavi2019generating}, patch embeddings from both CNN and transformer encoders can be represented with a finite token codebook, allowing for autoregressive image generation both unimodally~\cite{esser2021taming} and multimodally by conditioning on queries such as text descriptions of images~\cite{rombach2022high,yu2022scaling,ramesh2022hierarchical}.
Whether continuous or quantized, these formulations all encode images into standardized numbers of tokens, independent of image complexity or downstream task demands. 
In contrast, DOVE represents images using variable numbers of tokens, dynamically adapting to the complexity of images in unimodal settings and to the information demands of downstream tasks in text-conditioned ones.  

\textbf{Token Pruning and Compression.}
Token pruning methods reduce computation costs by iteratively reducing the set of tokens to be processed across transformer layers, either by dynamically omitting them~\cite{yin2022vit,rao2021dynamicvit} or by aggregating them in between layers of the transformer~\cite{bolya2023token}.
Because these methods iteratively modify the number of tokens across transformer layers, they require modification of the internal structure of models they are applied to. 
In contrast, DOVE produces variable numbers of tokens, allowing for it to be directly integrated into model pre-training and fine-tuning pipelines. 
Another branch of work reduces computational costs by compressing token sets at the input level. 
The Perceiver architecture uses a transformer to compress a set of input tokens into a smaller, fixed set of latent tokens~\cite{jaegle2021perceiver,jaegle2021perceiver_io}, allowing for greater computational tractability in multimodal settings~\cite{alayrac2022flamingo}.
Similarly, TiTok~\cite{yu2024image} compresses image patches into a small set of latent tokens, which are then quantized for image reconstruction or other downstream tasks. 

Closest to our work is ALIT~\cite{duggal2024adaptive}, which uses a recurrent process to distill 2D tokens into a set of 1D latent tokens. 
Although this iterative process allows for images to be represented by variable numbers of tokens, this is only evidenced through post-hoc analyses, and ALIT does not propose an automated method for dynamically determining the number of tokens to represent an image with at inference time. 
One of the key innovations of DOVE is the use of a dynamic EOS prediction mechanism, which is employed at inference time to produce per-image variable length token sequences based on image and downstream task complexity. DOVE uses a parallel transformer forward pass to generate variable number of tokens, which is more efficient ALIT's recurrent formulation. 

\textbf{Dynamic Sequence Termination.}
In the context of transformers, dynamic sequence termination is most commonly associated with the <EOS> token in LLMs~\cite{grattafiori2024llama,team2023gemini,achiam2023gpt}, although the concept has been applied in language modeling since N-gram models~\cite{chen1999empirical}.
This concept has also been generalized for generating variable length subsequences of specialized text, such as chain-of-thought chains generated between thinking tokens in LLMs~\cite{guo2025deepseek}.
In sequential decision making, dynamic termination has been operationalized through the use of terminal states in Hidden Markov Models~\cite{baum1966statistical}, termination conditions in the options reinforcement learning framework~\cite{sutton1999between}, as well as by using specialized stop actions within the low-level components of hierarchical policies~\cite{irshad2021hierarchical}.
\vspace{-5pt}

%% file: sec/5_conclusion.tex
\vspace{-3pt}
\section{Conclusion}
\vspace{-6pt}
We have introduced DOVE, a dynamic vision encoder that adaptively generates variable-length token sequences based on image complexity. DOVE predicts an end-of-sequence (EOS) token to dynamically determine the number of tokens needed for image reconstruction, resulting in significantly improved efficiency and semantic representation. We further extended our model with a query-conditioned variant, enabling task-specific focus on relevant image regions. Q-DOVE further improves the representations and token compression achieving stronger efficiency and performance. 

%% file: Appendix/Implementation.tex
\section{Implementation Details}

\subsection{Model Architecture}
Our framework builds on a pretrained VQGAN and two instances of the lightweight Pythia-70M language model. The VQGAN handles initial visual processing and image reconstruction, while two Pythia models are responsible for generating variable-length visual tokens and decoding them into a fixed-length sequence. Although our design uses transformer-based Pythia models to support dynamic sequence generation, the overall architecture remains lightweight, with a total parameter count roughly twice that of VQGAN alone. Details of the VQGAN and both Pythia-70M models we used are provided in Table~\ref{tab:model-details}.

\begin{table}[h]
\centering
\resizebox{\textwidth}{!}{%
\begin{tabular}{lcc}
\toprule
\textbf{Component} & \textbf{VQGAN} & \textbf{Pythia-70M} \\
\midrule
Type & Visual Tokenizer (Autoencoder) & Language Model (Decoder-only Transformer) \\
Parameters & $\sim$163M & $\sim$70M \\
Codebook Size & 8192 & -- \\
Embedding Dim & 256 & 512 \\
Layers & Encoder: 4, Decoder: 4 & 6 Transformer layers \\
Patch Size & 16$\times$16 & -- \\
Vocabulary Size & -- & 50,304 \\
Pretraining Dataset & ImageNet & The Pile \\
Usage in Our Framework & Image Tokenization and Reconstruction & Dynamic Token Generation and Decoding \\
\bottomrule
\end{tabular}
}
\vspace{2mm}
\caption{Model architecture details for VQGAN and Pythia-70M.}
\label{tab:model-details}
\end{table}

\subsection{Training Data}

We train our model on ImageNet-1K, a curated variant of the standard ImageNet dataset that contains 1.2 million images across 1,000 object categories. All images are resized to 256$\times$256, and data augmentation is applied using mild random cropping and grayscale adjustment to improve generalization. For Query-conditioned DOVE (Q-DOVE), we further fine-tune the original DOVE model on the Visual Genome and Open Images datasets for an additional five epochs. The Visual Genome dataset consists of 108,077 images, from which 5.4 million region descriptions and 1.7 million visual question–answer pairs are used as textual conditions. Additionally, we utilize 3.3 million relationship annotations from Open Images, where the bounding boxes of each object pair are spatially concatenated to define the conditioning region. Detailed statistics and usage of each dataset are summarized in Table~\ref{tab:training-data}.

\begin{table}[h]
\centering
\resizebox{\textwidth}{!}{%
\begin{tabular}{lccccc}
\toprule
\textbf{Dataset} & \textbf{\#Images} & \textbf{\#Textual Inputs} & \textbf{\#Relationship Annotations} & \textbf{Use Case} & \textbf{Epochs} \\
\midrule
ImageNet-1K     & 1.2M     & --                        & --                          & Pretraining (DOVE)       & 20 \\
Visual Genome   & 108K     & 5.4M region desc. + 1.7M QA & --                          & Fine-tuning (Q-DOVE) & 5 \\
Open Images     & 9M       & --                        & 3.3M relationships          & Fine-tuning (Q-DOVE) & 5 \\
\bottomrule
\end{tabular}
}
\vspace{2mm}
\caption{Training datasets used for DOVE and Q-DOVE. Textual inputs include region descriptions and question–answer pairs.}
\label{tab:training-data}
\end{table}

\subsection{Reconstruction Loss Function Design}

\begin{figure}[!h]
    \centering
    \includegraphics[width=\linewidth]{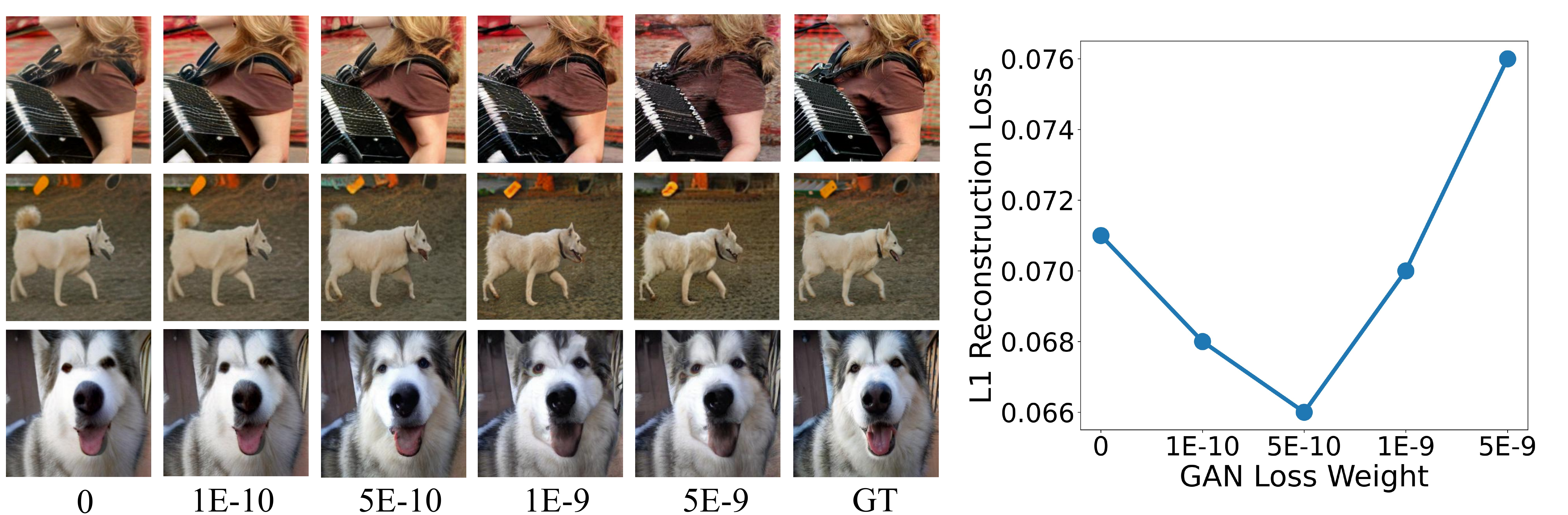}
    \caption{Effect of varying GAN loss weight on image reconstruction quality. A small weight (e.g., $5 \times 10^{-10}$) improves perceptual detail without sacrificing fidelity, while larger weights introduce artifacts and increase L1 loss.}
    \label{fig:gan-loss-comparison}
    \vspace{-12pt}
\end{figure}

To optimize image reconstruction, we combine mean squared error (MSE) loss, perceptual loss, and adversarial (GAN) loss. We find that incorporating a small weight for the GAN loss (e.g., $5 \times 10^{-10}$) enhances the realism and fine details of the reconstructed images. Figure~\ref{fig:gan-loss-comparison} presents some qualitative comparisons of reconstructions across a range of GAN loss weights, from 0 to $5 \times 10^{-9}$. As shown, increasing the GAN loss weight enhances texture detail; for example, the fur of a dog appears noticeably sharper with a weight of $5 \times 10^{-9}$ compared to reconstructions without GAN loss. However, assigning a larger GAN weight also introduces hallucinated content, leading to shape distortions and reduced fidelity to the original image. In addition, we evaluate the average L1 reconstruction loss on the ImageNet-1K validation set for each setting. The results indicate that a small GAN loss weight initially improves reconstruction accuracy. But when the weight increases further, the L1 loss also increases and eventually becomes higher than that of the model trained without GAN loss. Based on this trade-off, we choose $5 \times 10^{-10}$ as the GAN loss weight for our final model.

%% file: Appendix/VLM.tex
\section{Multimodal Understanding}

\subsection{Instruction Tuning Setup}

We follow the evaluation setup of AIM V2 and fine-tune Vicuna-7B-v1.5 models with different vision encoders on the 665K mixed VQA dataset from LLaVA. This mixed dataset includes training data from COCO, GQA, OCR-VQA, TextVQA, and Visual Genome. Detailed training configurations are provided in Table~\ref{tab:training-params}.

\begin{table}[h]
\centering
\resizebox{0.45\textwidth}{!}{%
\begin{tabular}{ll}
\toprule
\textbf{Config} & \textbf{LLaVA SFT Mixture} \\
\midrule
Optimizer & PyTorch's AdamW \\
Optimizer Momentum & $\beta_1 = 0.9,\ \beta_2 = 0.999$ \\
Decoder Peak Learning Rate & 2e-5 \\
Connector Peak Learning Rate & 2e-4 \\
Minimum Learning Rate & 0 \\
Weight Decay & 0.01 \\
Batch Size & 8 \\
Gradient Clipping & 1.0 \\
Warmup Iterations & 250 \\
Training Iterations & 5000 \\
Learning Rate Schedule & cosine decay \\
Transformations & [PadToSquare, Resize] \\
\bottomrule
\end{tabular}
}
\vspace{2mm}
\caption{Training configurations for fine-tuning VLM on the LLaVA SFT mixture.}
\label{tab:training-params}
\end{table}

\subsection{Evaluation Benchmarks}

We evaluate Vicuna models equipped with different vision encoders across eight diverse datasets. Table~\ref{tab:evaluation-benchmarks} summarizes the benchmarks used in our evaluation, including dataset split, prompt style, and evaluation metric.

\begin{table}[h]
\centering
\resizebox{\textwidth}{!}{%
\begin{tabular}{llll}
\toprule
\textbf{Benchmark} & \textbf{Split} & \textbf{Prompt} & \textbf{Evaluation Metric} \\
\midrule
VQAv2~\cite{goyal2017makingvvqamatter} & Val &  & Accuracy \\
GQA~\cite{ainslie2023gqatraininggeneralizedmultiquery} & Val &  & Accuracy \\
OK-VQA~\cite{marino2019okvqavisualquestionanswering} & Val &  & Accuracy \\
TextVQA~\cite{singh2019vqamodelsread} & Val & \textit{Answer the question using a single word or phrase.} & Accuracy \\
DocVQA~\cite{mathew2021docvqadatasetvqadocument} & Test &  & ANLS \\
InfoVQA~\cite{mathew2021infographicvqa} & Test &  & ANLS \\
ChartQA~\cite{masry2022chartqabenchmarkquestionanswering} & Test &  & Relaxed Accuracy \\
ScienceQA~\cite{lu2022learnexplainmultimodalreasoning} & Test (image split) & \textit{Answer with the option’s letter from the given choices directly.} & Accuracy \\
\bottomrule
\end{tabular}
}
\vspace{2mm}
\caption{Evaluation benchmarks used in Visual Question Answering Evaluation.}
\label{tab:evaluation-benchmarks}
\end{table}

\subsection{Case Study}

\begin{figure}[!h]
    \centering
    \includegraphics[width=\linewidth]{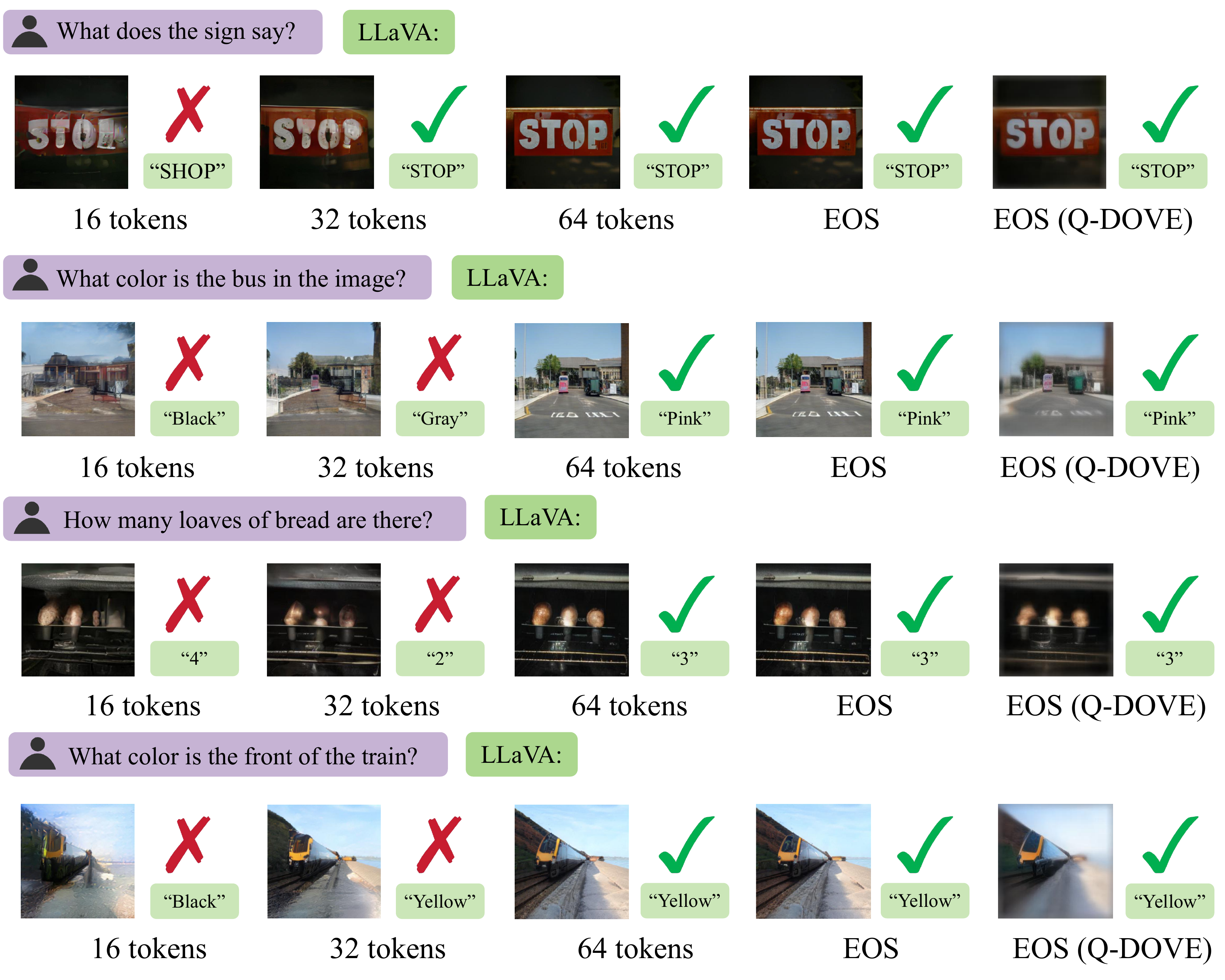}
    \caption{Model predictions under varying token counts. As the number of tokens increases, both image reconstruction quality and answer accuracy improve.}
    \label{fig:case_study}
    \vspace{-12pt}
\end{figure}

We conduct a case study to analyze the VLM's responses under different token counts. Figure~\ref{fig:case_study} shows reconstructed images and the corresponding answers generated by the model. We find that as the number of tokens increases, both reconstructed image quality and answer accuracy improve. With fewer tokens, the images become blurry and the VLM is more likely to hallucinate; for example, when using only 16 tokens, the VLM misreads the word ``STOP'' on a sign as ``SHOP''.

%% file: Appendix/Linear_Prob.tex
\section{Linear Probing Datasets}

We evaluate the quality of visual representations in vision encoders using a standard linear probing setup. In this setup, the vision encoder is frozen, and a single-layer linear classifier is trained on top of a selected hidden layer to perform image classification. The classifier is trained and evaluated across a suite of standard benchmarks, including CIFAR-10~\cite{krizhevsky2009learning}, CIFAR-100~\cite{krizhevsky2009learning}, DTD~\cite{cimpoi14describing}, FGVC-Aircraft~\cite{maji2013finegrainedvisualclassificationaircraft}, Food101~\cite{bossard14}, STL-10~\cite{coates2011analysis}, and SUN397~\cite{5539970}. Descriptions of each dataset are provided below.

\paragraph{CIFAR-10 / CIFAR-100} 
CIFAR-10 and CIFAR-100~\cite{krizhevsky2009learning} are widely used benchmarks for object classification. Both datasets consist of 60,000 low-resolution (32$\times$32) color images, with 50,000 for training and 10,000 for testing. CIFAR-10 includes 10 coarse classes such as airplane, dog, and truck, while CIFAR-100 contains 100 fine-grained categories organized into 20 superclasses.

\paragraph{DTD} 
The Describable Textures Dataset (DTD)~\cite{cimpoi14describing} contains 5,640 images organized into 47 texture categories annotated with human-interpretable attributes (e.g., “striped”, “dotted”, “bumpy”). The images are collected from the wild, with significant variation in lighting, scale, and viewpoint, serving as a benchmark for texture recognition and attribute prediction.

\paragraph{FGVC-Aircraft} 
FGVC-Aircraft~\cite{maji2013finegrainedvisualclassificationaircraft} is a fine-grained classification dataset containing 10,000 images of aircraft across 100 classes, such as Boeing 747 and Airbus A320. The dataset emphasizes subtle inter-class visual differences, with consistent poses but variations in background and lighting.

\paragraph{Food101}
Food101~\cite{bossard14} comprises 101,000 high-resolution images across 101 food categories, with 1,000 images per class. The dataset reflects real-world variability, including occlusions, diverse lighting conditions, and a broad range of cuisines and presentation styles.

\paragraph{STL-10}
STL-10~\cite{coates2011analysis} is designed for unsupervised and semi-supervised learning. It includes 13,000 labeled images spanning 10 object categories (e.g., bird, cat, ship) at a resolution of 96$\times$96, along with an additional 100,000 unlabeled images for representation learning.

\paragraph{SUN397}
SUN397~\cite{5539970} is a large-scale scene classification dataset comprising over 100,000 images across 397 categories. It includes a diverse range of indoor and outdoor environments such as kitchens, libraries, highways, and mountains. The dataset is intended to assess a model’s ability to recognize complex and varied semantic scenes.

%% file: Appendix/Quantize.tex
\section{Experiments with Gaussian Latent Space}
\begin{figure}[!h]
    \centering
    \includegraphics[width=\linewidth]{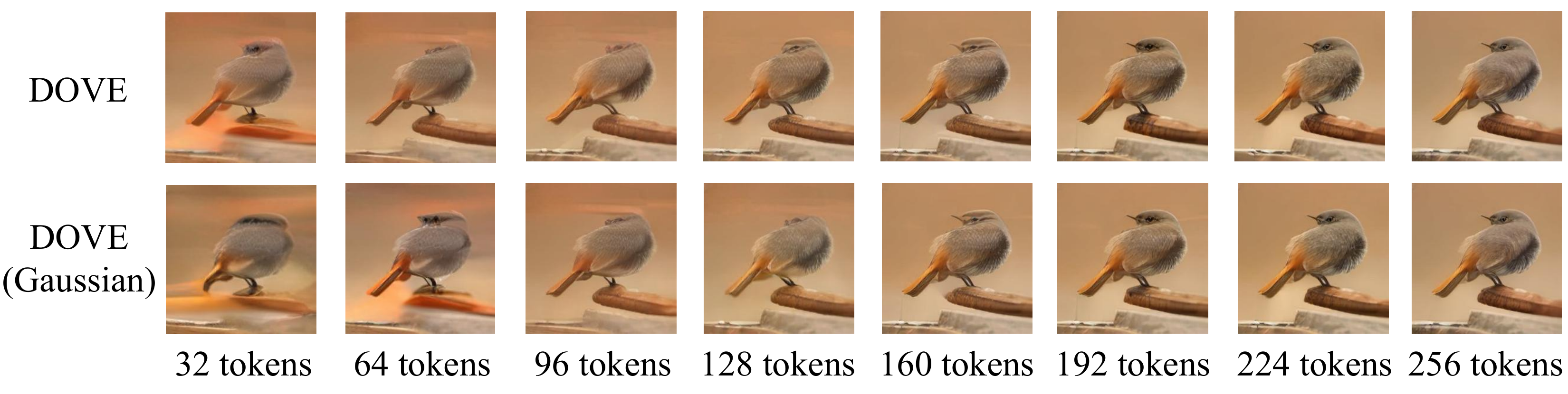}
    \caption{Reconstruction results of DOVE and DOVE (Gaussian) under varying token budgets. Overall, DOVE (Gaussian) achieves similar visual quality to DOVE.}
    \label{fig:Gaussain}
    \vspace{-12pt}
\end{figure}

To ensure that the generated representations converge to a known distribution, we adopt the reparameterization technique from variational autoencoders (VAEs). Specifically, we map the tokens generated by DOVE into a Gaussian latent space. Our results show that after Gaussianization, the model maintains a reconstruction quality comparable to the original version. FID scores are reported in Table~\ref{tab:Gaussain}, and qualitative examples are shown in Figure~\ref{fig:Gaussain}.

We also observe that the token representations generated by DOVE are unevenly distributed. For example, most of the information is concentrated in the first 64 tokens, while the remaining tokens contribute only subtle variations. This uneven distribution poses challenges for effective quantization into a discrete representation space such as a codebook. We will further investigate improved quantization strategies for DOVE in future work.

\begin{table*}[ht]
\centering
\resizebox{\textwidth}{!}{%
\begin{tabular}{lcccccccccccccc}
\toprule
\multirow{2}{*}{\textbf{Approach}} 
& \multicolumn{8}{c}{\textbf{ImageNet100}} 
& \multicolumn{3}{c}{\textbf{COCO}} 
& \multicolumn{3}{c}{\textbf{Wikipedia (WIT)}} \\
\cmidrule(lr){2-9} \cmidrule(lr){10-12} \cmidrule(lr){13-15}
 & 32 & 64 & 96 & 128 & 160 & 192 & 224 & 256 
 & 32$^\#$ / 64 & 128 & 256 
 & 32$^\#$ / 64 & 128 & 256 \\
\midrule
TiTok-L-32      & 11.60 & -    & -    & -    & -    & -    & -    & -    & 14.18$^\#$ & -     & -     & 53.57$^\#$ & -     & -     \\
TiTok-B-64      & -     & 8.22 & -    & -    & -    & -    & -    & -    & 9.15       & -     & -     & 42.86      & -     & -     \\
TiTok-S-128     & -     & -    & - & 8.22    & -    & -    & -    & -    & -          & 9.15     & -     & -          & 38.16 & -     \\
VQGAN          & -     & -    & -    & -    & -    & -    & -    & 7.04    & -       & -     & 7.77     & -          & -     & 31.27  \\
ALIT          & 22.31 & 15.92 & 13.08 & 11.45 & 10.01 & 9.12 & 8.37 & 8.06 & 22.01 & 13.98 & 9.51 & 61.32 & 47.52 & 38.10 \\
DOVE  & 18.91 & 11.46 & 10.84 & 9.28 &  8.61 & 8.25 & 7.96 & 7.73 & 15.50 & 9.83 & 7.54 & 14.83 & 8.56 & 7.84      \\
DOVE (Gaussian)  & 19.87 & 12.03 & 10.98 & 9.46 &  8.95 & 8.28 & 8.01 & 7.80 & 16.12 & 10.03 & 7.58 & 15.34 & 8.84 & 7.87      \\
\bottomrule
\end{tabular}%
}
\caption{FID comparison between DOVE and DOVE (Gaussian) on various datasets.}
\label{tab:Gaussain}
\end{table*}

%% file: main.bbl
\begin{thebibliography}{10}

\bibitem{achiam2023gpt}
Josh Achiam, Steven Adler, Sandhini Agarwal, Lama Ahmad, Ilge Akkaya, Florencia~Leoni Aleman, Diogo Almeida, Janko Altenschmidt, Sam Altman, Shyamal Anadkat, et~al.
\newblock Gpt-4 technical report.
\newblock {\em arXiv preprint arXiv:2303.08774}, 2023.

\bibitem{ainslie2023gqatraininggeneralizedmultiquery}
Joshua Ainslie, James Lee-Thorp, Michiel de~Jong, Yury Zemlyanskiy, Federico Lebrón, and Sumit Sanghai.
\newblock Gqa: Training generalized multi-query transformer models from multi-head checkpoints, 2023.

\bibitem{alayrac2022flamingo}
Jean-Baptiste Alayrac, Jeff Donahue, Pauline Luc, Antoine Miech, Iain Barr, Yana Hasson, Karel Lenc, Arthur Mensch, Katherine Millican, Malcolm Reynolds, et~al.
\newblock Flamingo: a visual language model for few-shot learning.
\newblock {\em Advances in neural information processing systems}, 35:23716--23736, 2022.

\bibitem{bajcsy2018revisiting}
Ruzena Bajcsy, Yiannis Aloimonos, and John~K Tsotsos.
\newblock Revisiting active perception.
\newblock {\em Autonomous Robots}, 42:177--196, 2018.

\bibitem{bansal2016blur}
Raghav Bansal, Gaurav Raj, and Tanupriya Choudhury.
\newblock Blur image detection using laplacian operator and open-cv.
\newblock In {\em 2016 International Conference System Modeling \& Advancement in Research Trends (SMART)}, pages 63--67. IEEE, 2016.

\bibitem{baum1966statistical}
Leonard~E Baum and Ted Petrie.
\newblock Statistical inference for probabilistic functions of finite state markov chains.
\newblock {\em The annals of mathematical statistics}, 37(6):1554--1563, 1966.

\bibitem{biderman2023pythia}
Stella Biderman, Hailey Schoelkopf, Quentin~Gregory Anthony, Herbie Bradley, Kyle O’Brien, Eric Hallahan, Mohammad~Aflah Khan, Shivanshu Purohit, USVSN~Sai Prashanth, Edward Raff, et~al.
\newblock Pythia: A suite for analyzing large language models across training and scaling.
\newblock In {\em International Conference on Machine Learning}, pages 2397--2430. PMLR, 2023.

\bibitem{bolya2023token}
Daniel Bolya, Cheng-Yang Fu, Xiaoliang Dai, Peizhao Zhang, Christoph Feichtenhofer, and Judy Hoffman.
\newblock Token merging: Your vit but faster.
\newblock In {\em The Eleventh International Conference on Learning Representations}, 2023.

\bibitem{bossard14}
Lukas Bossard, Matthieu Guillaumin, and Luc Van~Gool.
\newblock Food-101 -- mining discriminative components with random forests.
\newblock In {\em European Conference on Computer Vision}, 2014.

\bibitem{caron2021emerging}
Mathilde Caron, Hugo Touvron, Ishan Misra, Herv{\'e} J{\'e}gou, Julien Mairal, Piotr Bojanowski, and Armand Joulin.
\newblock Emerging properties in self-supervised vision transformers.
\newblock In {\em Proceedings of the IEEE/CVF international conference on computer vision}, pages 9650--9660, 2021.

\bibitem{chen2024efficient}
Jieneng Chen, Luoxin Ye, Ju~He, Zhao-Yang Wang, Daniel Khashabi, and Alan Yuille.
\newblock Efficient large multi-modal models via visual context compression.
\newblock In {\em The Thirty-eighth Annual Conference on Neural Information Processing Systems}, 2024.

\bibitem{chen2021review}
Leiyu Chen, Shaobo Li, Qiang Bai, Jing Yang, Sanlong Jiang, and Yanming Miao.
\newblock Review of image classification algorithms based on convolutional neural networks.
\newblock {\em Remote Sensing}, 13(22):4712, 2021.

\bibitem{chen1999empirical}
Stanley~F Chen and Joshua Goodman.
\newblock An empirical study of smoothing techniques for language modeling.
\newblock {\em Computer Speech \& Language}, 13(4):359--394, 1999.

\bibitem{cimpoi14describing}
M.~Cimpoi, S.~Maji, I.~Kokkinos, S.~Mohamed, , and A.~Vedaldi.
\newblock Describing textures in the wild.
\newblock In {\em Proceedings of the {IEEE} Conf. on Computer Vision and Pattern Recognition ({CVPR})}, 2014.

\bibitem{coates2011analysis}
Adam Coates, Honglak Lee, and AY~Ng.
\newblock An analysis of single layer networks in unsupervised feature learning aistats.
\newblock 2011.

\bibitem{cui2023scaling}
Justin Cui, Ruochen Wang, Si~Si, and Cho-Jui Hsieh.
\newblock Scaling up dataset distillation to imagenet-1k with constant memory.
\newblock In {\em International Conference on Machine Learning}, pages 6565--6590. PMLR, 2023.

\bibitem{deangelus2009top}
Marianne DeAngelus and Jeff~B Pelz.
\newblock Top-down control of eye movements: Yarbus revisited.
\newblock {\em Visual Cognition}, 17(6-7):790--811, 2009.

\bibitem{deng2009imagenet}
Jia Deng, Wei Dong, Richard Socher, Li-Jia Li, Kai Li, and Li~Fei-Fei.
\newblock Imagenet: A large-scale hierarchical image database.
\newblock In {\em 2009 IEEE conference on computer vision and pattern recognition}, pages 248--255. Ieee, 2009.

\bibitem{dosovitskiy2021an}
Alexey Dosovitskiy, Lucas Beyer, Alexander Kolesnikov, Dirk Weissenborn, Xiaohua Zhai, Thomas Unterthiner, Mostafa Dehghani, Matthias Minderer, Georg Heigold, Sylvain Gelly, Jakob Uszkoreit, and Neil Houlsby.
\newblock An image is worth 16x16 words: Transformers for image recognition at scale.
\newblock In {\em International Conference on Learning Representations}, 2021.

\bibitem{duggal2024adaptive}
Shivam Duggal, Phillip Isola, Antonio Torralba, and William~T Freeman.
\newblock Adaptive length image tokenization via recurrent allocation.
\newblock In {\em First Workshop on Scalable Optimization for Efficient and Adaptive Foundation Models}, 2024.

\bibitem{esser2021taming}
Patrick Esser, Robin Rombach, and Bjorn Ommer.
\newblock Taming transformers for high-resolution image synthesis.
\newblock In {\em Proceedings of the IEEE/CVF conference on computer vision and pattern recognition}, pages 12873--12883, 2021.

\bibitem{fini_multimodal-autoregressive}
Enrico Fini*, Mustafa Shukor*, Xiujun Li, Philipp Dufter, Michal Klein, David Haldimann, Sai Aitharaju, Louis Béthune, Zhe Gan, Victor Turrisi, Alexander Toshev, Marcin Eichner, Yinfei Yang, Moin Nabi, Josh Susskind, and Alaaeldin El-Nouby*.
\newblock Multimodal autoregressive pre-training of large vision encoders, 2024.

\bibitem{goyal2017makingvvqamatter}
Yash Goyal, Tejas Khot, Douglas Summers-Stay, Dhruv Batra, and Devi Parikh.
\newblock Making the v in vqa matter: Elevating the role of image understanding in visual question answering, 2017.

\bibitem{grattafiori2024llama}
Aaron Grattafiori, Abhimanyu Dubey, Abhinav Jauhri, Abhinav Pandey, Abhishek Kadian, Ahmad Al-Dahle, Aiesha Letman, Akhil Mathur, Alan Schelten, Alex Vaughan, et~al.
\newblock The llama 3 herd of models.
\newblock {\em arXiv preprint arXiv:2407.21783}, 2024.

\bibitem{guo2025deepseek}
Daya Guo, Dejian Yang, Haowei Zhang, Junxiao Song, Ruoyu Zhang, Runxin Xu, Qihao Zhu, Shirong Ma, Peiyi Wang, Xiao Bi, et~al.
\newblock Deepseek-r1: Incentivizing reasoning capability in llms via reinforcement learning.
\newblock {\em arXiv preprint arXiv:2501.12948}, 2025.

\bibitem{guo2018review}
Yanming Guo, Yu~Liu, Theodoros Georgiou, and Michael~S Lew.
\newblock A review of semantic segmentation using deep neural networks.
\newblock {\em International journal of multimedia information retrieval}, 7:87--93, 2018.

\bibitem{hao2020brief}
Shijie Hao, Yuan Zhou, and Yanrong Guo.
\newblock A brief survey on semantic segmentation with deep learning.
\newblock {\em Neurocomputing}, 406:302--321, 2020.

\bibitem{irshad2021hierarchical}
Muhammad~Zubair Irshad, Chih-Yao Ma, and Zsolt Kira.
\newblock Hierarchical cross-modal agent for robotics vision-and-language navigation.
\newblock In {\em 2021 IEEE international conference on robotics and automation (ICRA)}, pages 13238--13246. IEEE, 2021.

\bibitem{jaegle2021perceiver_io}
Andrew Jaegle, Sebastian Borgeaud, Jean-Baptiste Alayrac, Carl Doersch, Catalin Ionescu, David Ding, Skanda Koppula, Daniel Zoran, Andrew Brock, Evan Shelhamer, et~al.
\newblock Perceiver io: A general architecture for structured inputs \& outputs.
\newblock {\em arXiv preprint arXiv:2107.14795}, 2021.

\bibitem{jaegle2021perceiver}
Andrew Jaegle, Felix Gimeno, Andy Brock, Oriol Vinyals, Andrew Zisserman, and Joao Carreira.
\newblock Perceiver: General perception with iterative attention.
\newblock In {\em International conference on machine learning}, pages 4651--4664. PMLR, 2021.

\bibitem{kingma2013auto}
Diederik~P Kingma, Max Welling, et~al.
\newblock Auto-encoding variational bayes, 2013.

\bibitem{krishna2017visual}
Ranjay Krishna, Yuke Zhu, Oliver Groth, Justin Johnson, Kenji Hata, Joshua Kravitz, Stephanie Chen, Yannis Kalantidis, Li-Jia Li, David~A Shamma, et~al.
\newblock Visual genome: Connecting language and vision using crowdsourced dense image annotations.
\newblock {\em International journal of computer vision}, 123:32--73, 2017.

\bibitem{krizhevsky2009learning}
Alex Krizhevsky, Geoffrey Hinton, et~al.
\newblock Learning multiple layers of features from tiny images.
\newblock 2009.

\bibitem{kuznetsova2020open}
Alina Kuznetsova, Hassan Rom, Neil Alldrin, Jasper Uijlings, Ivan Krasin, Jordi Pont-Tuset, Shahab Kamali, Stefan Popov, Matteo Malloci, Alexander Kolesnikov, et~al.
\newblock The open images dataset v4: Unified image classification, object detection, and visual relationship detection at scale.
\newblock {\em International journal of computer vision}, 128(7):1956--1981, 2020.

\bibitem{land1999roles}
Michael Land, Neil Mennie, and Jennifer Rusted.
\newblock The roles of vision and eye movements in the control of activities of daily living.
\newblock {\em Perception}, 28(11):1311--1328, 1999.

\bibitem{lin2014microsoft}
Tsung-Yi Lin, Michael Maire, Serge Belongie, James Hays, Pietro Perona, Deva Ramanan, Piotr Doll{\'a}r, and C~Lawrence Zitnick.
\newblock Microsoft coco: Common objects in context.
\newblock In {\em Computer vision--ECCV 2014: 13th European conference, zurich, Switzerland, September 6-12, 2014, proceedings, part v 13}, pages 740--755. Springer, 2014.

\bibitem{liu2023llava}
Haotian Liu, Chunyuan Li, Qingyang Wu, and Yong~Jae Lee.
\newblock Visual instruction tuning, 2023.

\bibitem{lu2007survey}
Dengsheng Lu and Qihao Weng.
\newblock A survey of image classification methods and techniques for improving classification performance.
\newblock {\em International journal of Remote sensing}, 28(5):823--870, 2007.

\bibitem{lu2022learnexplainmultimodalreasoning}
Pan Lu, Swaroop Mishra, Tony Xia, Liang Qiu, Kai-Wei Chang, Song-Chun Zhu, Oyvind Tafjord, Peter Clark, and Ashwin Kalyan.
\newblock Learn to explain: Multimodal reasoning via thought chains for science question answering, 2022.

\bibitem{maji2013finegrainedvisualclassificationaircraft}
Subhransu Maji, Esa Rahtu, Juho Kannala, Matthew Blaschko, and Andrea Vedaldi.
\newblock Fine-grained visual classification of aircraft, 2013.

\bibitem{marino2019okvqavisualquestionanswering}
Kenneth Marino, Mohammad Rastegari, Ali Farhadi, and Roozbeh Mottaghi.
\newblock Ok-vqa: A visual question answering benchmark requiring external knowledge, 2019.

\bibitem{masry2022chartqabenchmarkquestionanswering}
Ahmed Masry, Do~Xuan Long, Jia~Qing Tan, Shafiq Joty, and Enamul Hoque.
\newblock Chartqa: A benchmark for question answering about charts with visual and logical reasoning, 2022.

\bibitem{mathew2021infographicvqa}
Minesh Mathew, Viraj Bagal, Rubèn~Pérez Tito, Dimosthenis Karatzas, Ernest Valveny, and C.~V Jawahar.
\newblock Infographicvqa, 2021.

\bibitem{mathew2021docvqadatasetvqadocument}
Minesh Mathew, Dimosthenis Karatzas, and C.~V. Jawahar.
\newblock Docvqa: A dataset for vqa on document images, 2021.

\bibitem{N/A_2024}
N/A.
\newblock Stl-10, nov 2024.

\bibitem{radford2021learning}
Alec Radford, Jong~Wook Kim, Chris Hallacy, Aditya Ramesh, Gabriel Goh, Sandhini Agarwal, Girish Sastry, Amanda Askell, Pamela Mishkin, Jack Clark, et~al.
\newblock Learning transferable visual models from natural language supervision.
\newblock In {\em International conference on machine learning}, pages 8748--8763. PmLR, 2021.

\bibitem{ramesh2022hierarchical}
Aditya Ramesh, Prafulla Dhariwal, Alex Nichol, Casey Chu, and Mark Chen.
\newblock Hierarchical text-conditional image generation with clip latents.
\newblock {\em arXiv preprint arXiv:2204.06125}, 1(2):3, 2022.

\bibitem{rao2021dynamicvit}
Yongming Rao, Wenliang Zhao, Benlin Liu, Jiwen Lu, Jie Zhou, and Cho-Jui Hsieh.
\newblock Dynamicvit: Efficient vision transformers with dynamic token sparsification.
\newblock {\em Advances in neural information processing systems}, 34:13937--13949, 2021.

\bibitem{razavi2019generating}
Ali Razavi, Aaron Van~den Oord, and Oriol Vinyals.
\newblock Generating diverse high-fidelity images with vq-vae-2.
\newblock {\em Advances in neural information processing systems}, 32, 2019.

\bibitem{rombach2022high}
Robin Rombach, Andreas Blattmann, Dominik Lorenz, Patrick Esser, and Bj{\"o}rn Ommer.
\newblock High-resolution image synthesis with latent diffusion models.
\newblock In {\em Proceedings of the IEEE/CVF conference on computer vision and pattern recognition}, pages 10684--10695, 2022.

\bibitem{singh2019vqamodelsread}
Amanpreet Singh, Vivek Natarajan, Meet Shah, Yu~Jiang, Xinlei Chen, Dhruv Batra, Devi Parikh, and Marcus Rohrbach.
\newblock Towards vqa models that can read, 2019.

\bibitem{sutton1999between}
Richard~S Sutton, Doina Precup, and Satinder Singh.
\newblock Between mdps and semi-mdps: A framework for temporal abstraction in reinforcement learning.
\newblock {\em Artificial intelligence}, 112(1-2):181--211, 1999.

\bibitem{team2023gemini}
Gemini Team, Rohan Anil, Sebastian Borgeaud, Jean-Baptiste Alayrac, Jiahui Yu, Radu Soricut, Johan Schalkwyk, Andrew~M Dai, Anja Hauth, Katie Millican, et~al.
\newblock Gemini: a family of highly capable multimodal models.
\newblock {\em arXiv preprint arXiv:2312.11805}, 2023.

\bibitem{van2017neural}
Aaron Van Den~Oord, Oriol Vinyals, et~al.
\newblock Neural discrete representation learning.
\newblock {\em Advances in neural information processing systems}, 30, 2017.

\bibitem{vaswani_attn}
Ashish Vaswani, Noam Shazeer, Niki Parmar, Jakob Uszkoreit, Llion Jones, Aidan~N Gomez, \L~ukasz Kaiser, and Illia Polosukhin.
\newblock Attention is all you need.
\newblock In I.~Guyon, U.~Von Luxburg, S.~Bengio, H.~Wallach, R.~Fergus, S.~Vishwanathan, and R.~Garnett, editors, {\em Advances in Neural Information Processing Systems}, volume~30. Curran Associates, Inc., 2017.

\bibitem{xia2014supervised}
Rongkai Xia, Yan Pan, Hanjiang Lai, Cong Liu, and Shuicheng Yan.
\newblock Supervised hashing for image retrieval via image representation learning.
\newblock In {\em Proceedings of the AAAI conference on artificial intelligence}, volume~28, 2014.

\bibitem{5539970}
Jianxiong Xiao, James Hays, Krista~A. Ehinger, Aude Oliva, and Antonio Torralba.
\newblock Sun database: Large-scale scene recognition from abbey to zoo.
\newblock In {\em 2010 IEEE Computer Society Conference on Computer Vision and Pattern Recognition}, pages 3485--3492, 2010.

\bibitem{yin2022vit}
Hongxu Yin, Arash Vahdat, Jose~M Alvarez, Arun Mallya, Jan Kautz, and Pavlo Molchanov.
\newblock A-vit: Adaptive tokens for efficient vision transformer.
\newblock In {\em Proceedings of the IEEE/CVF conference on computer vision and pattern recognition}, pages 10809--10818, 2022.

\bibitem{yu2022scaling}
Jiahui Yu, Yuanzhong Xu, Jing~Yu Koh, Thang Luong, Gunjan Baid, Zirui Wang, Vijay Vasudevan, Alexander Ku, Yinfei Yang, Burcu~Karagol Ayan, et~al.
\newblock Scaling autoregressive models for content-rich text-to-image generation.
\newblock {\em arXiv preprint arXiv:2206.10789}, 2(3):5, 2022.

\bibitem{yu2024image}
Qihang Yu, Mark Weber, Xueqing Deng, Xiaohui Shen, Daniel Cremers, and Liang-Chieh Chen.
\newblock An image is worth 32 tokens for reconstruction and generation.
\newblock {\em Advances in Neural Information Processing Systems}, 37:128940--128966, 2024.

\bibitem{zhao2019object}
Zhong-Qiu Zhao, Peng Zheng, Shou-tao Xu, and Xindong Wu.
\newblock Object detection with deep learning: A review.
\newblock {\em IEEE transactions on neural networks and learning systems}, 30(11):3212--3232, 2019.

\bibitem{zou2023object}
Zhengxia Zou, Keyan Chen, Zhenwei Shi, Yuhong Guo, and Jieping Ye.
\newblock Object detection in 20 years: A survey.
\newblock {\em Proceedings of the IEEE}, 111(3):257--276, 2023.

\end{thebibliography}
